\documentclass{article}

\usepackage[final]{neurips_2025}
\usepackage{makecell}

\usepackage[utf8]{inputenc} 
\usepackage[T1]{fontenc}    
\usepackage{hyperref}       
\usepackage{url}            
\usepackage{booktabs}       
\usepackage{amsfonts}       
\usepackage{amsmath}   
\usepackage{cleveref}  
\usepackage{nicefrac}       
\usepackage{microtype}      
\usepackage{tcolorbox}
\usepackage{minitoc}
\usepackage[table,xcdraw]{xcolor} 
\usepackage{xspace}     
\usepackage{wrapfig}
\usepackage{float}
\usepackage{subcaption}
\usepackage{bbm}
\usepackage{multirow}

\definecolor{my_green}{RGB}{46, 151, 78}
\definecolor{my_blue}{RGB}{45, 125, 187}
\definecolor{my_red}{RGB}{216, 37, 34}
\newcommand{\PPercep}{\textbf{\textcolor{my_red}{P'Percep}}}
\newcommand{\PReason}{\textbf{\textcolor{my_blue}{P'Reason}}}
\newcommand{\PRobust}{\textbf{\textcolor{my_green}{P'Robust}}}

\newcommand{\VPP}{\textbf{\textcolor{my_red}{VPP}}} 
\newcommand{\CLP}{\textbf{\textcolor{my_red}{CLP}}} 
\newcommand{\LDP}{\textbf{\textcolor{my_red}{LDP}}} 
\newcommand{\VAP}{\textbf{\textcolor{my_red}{VAP}}} 

\newcommand{\PTR}{\textbf{\textcolor{my_blue}{PTR}}} 
\newcommand{\LRR}{\textbf{\textcolor{my_blue}{LRR}}} 
\newcommand{\PTS}{\textbf{\textcolor{my_blue}{PTS}}} 
\newcommand{\VPC}{\textbf{\textcolor{my_blue}{VPC}}} 
\newcommand{\OVR}{\textbf{\textcolor{my_blue}{OVR}}} 
\newcommand{\NAME}{{MMPerspective}\xspace}
\def\logo{\makebox[0pt][l]{\hspace{0pt}\raisebox{-0.7ex}{\includegraphics[height=24pt]{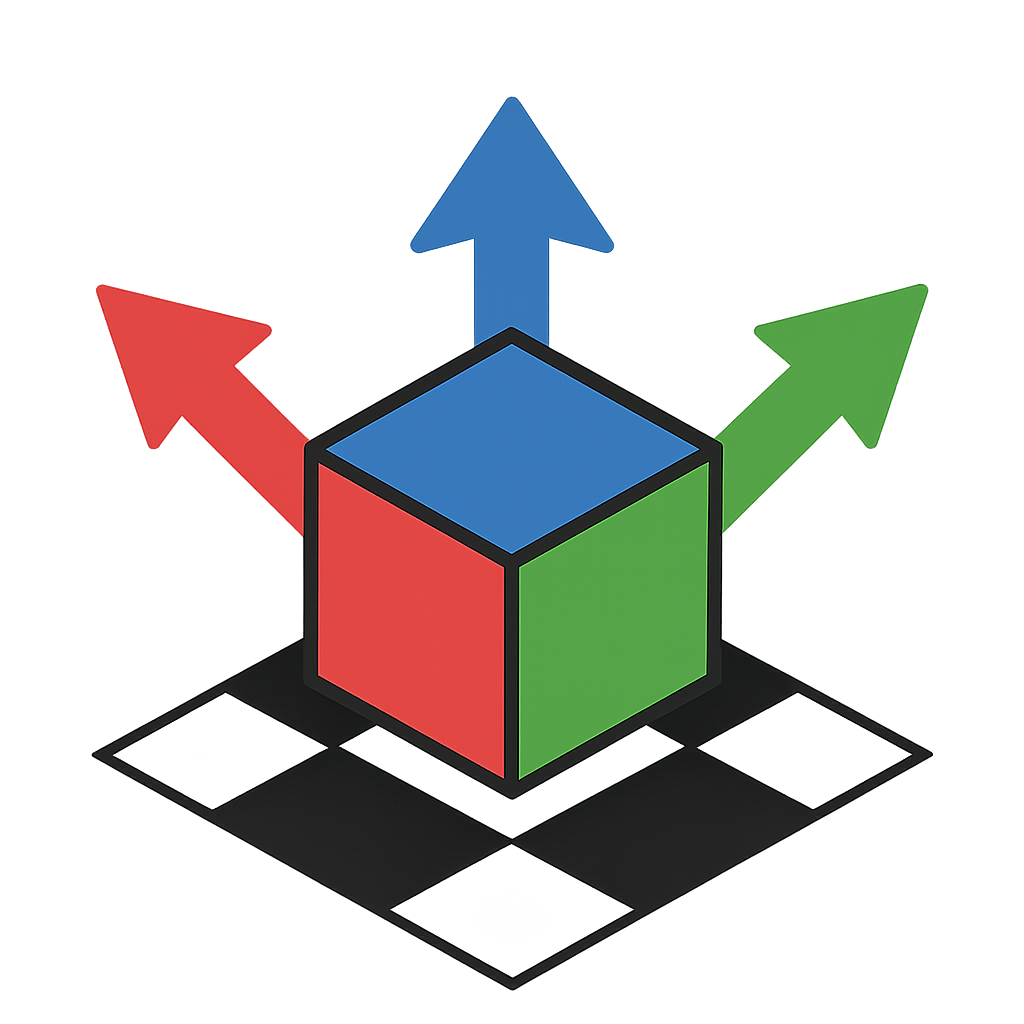}}}}

\title{\logo \ \ \ \ \ \ \NAME: Do MLLMs Understand Perspective? A Comprehensive Benchmark for Perspective Perception, Reasoning, and Robustness}

\author{
Yolo Yunlong Tang$^{1,*}$, Pinxin Liu$^{1,*}$, Zhangyun Tan$^{1,}\thanks{Equal contribution.}$,~~Mingqian Feng$^{1}$,
Rui Mao$^{1}$, \\ \textbf{Chao Huang$^{1}$,} \textbf{Jing Bi$^{1}$}, \textbf{Yunzhong Xiao$^{2}$,}
\textbf{Susan Liang$^{1}$,} \textbf{Hang Hua$^{1}$,}\\ \textbf{Ali Vosoughi$^{1}$,} \textbf{Luchuan Song$^{1}$,} \textbf{Zeliang Zhang$^{1}$,} \textbf{Chenliang Xu}$^{1}$ \\ \\
$^{1}$University of Rochester, $^{2}$Carnegie Mellon University \\ \\
{\tt\small \{yunlong.tang, mingqian.feng, jing.bi, chenliang.xu\}@rochester.edu}, \\
{\tt\small \{pliu23, rmao6, lsong11, zzh136\}@ur.rochester.edu},\\
{\tt\small ztan12@u.rochester.edu, \{chuang65, sliang22, hhua2\}@cs.rochester.edu}, \\
{\tt\small avosoughi@ece.rochester.edu, 
yunzhonx@andrew.cmu.edu}\\
}

\begin{document}
\maketitle

\begin{figure}[H]
\vspace{-2em}
    \centering
    \includegraphics[width=1\linewidth]{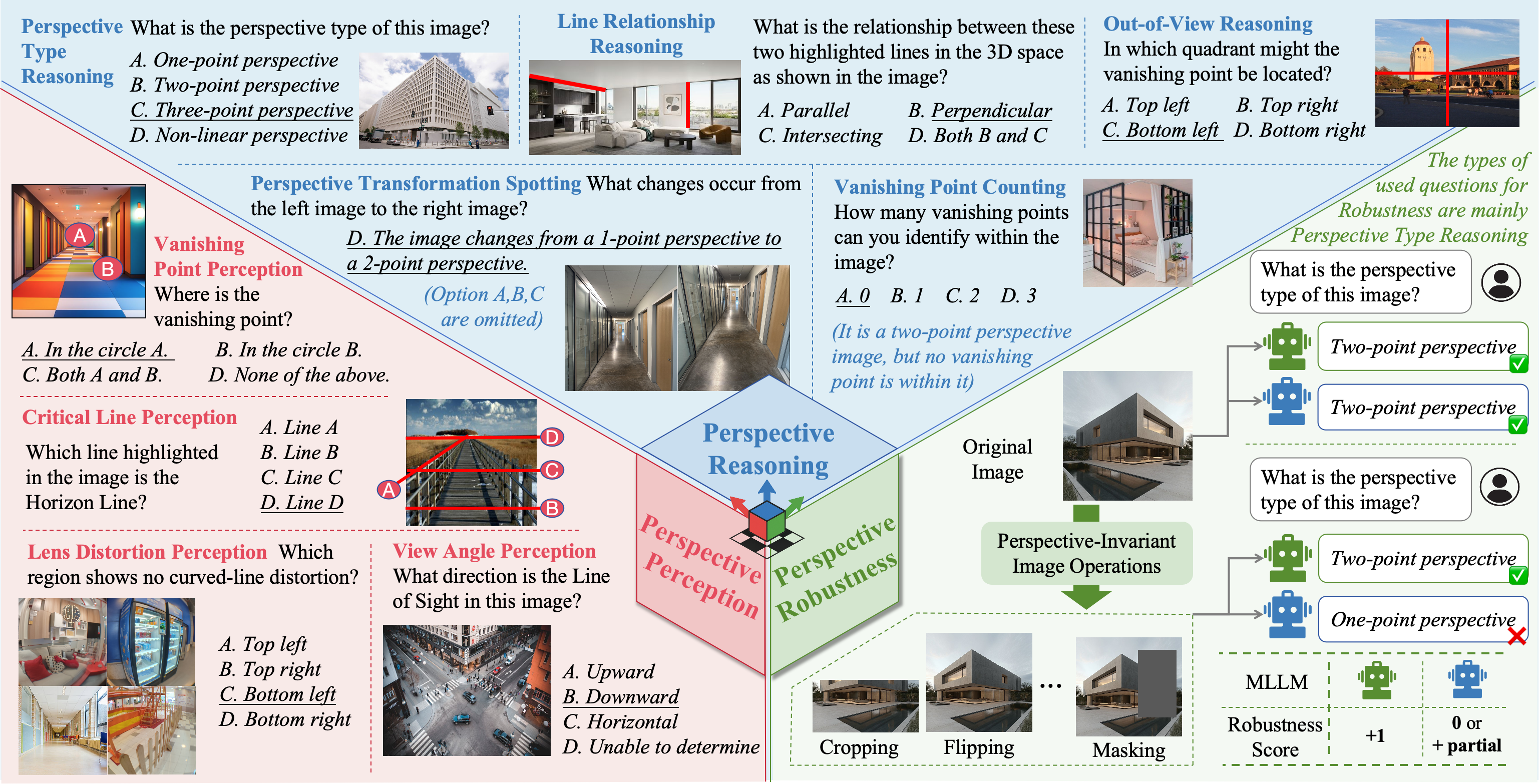}
    \caption{\textbf{MMPerspective benchmark overview.} We introduce 10 tasks spanning 3 complementary dimensions of perspective understanding: Perspective \textcolor{my_red}{\textbf{Perception}}, \textcolor{my_blue}{\textbf{Reasoning}}, and \textcolor{my_green}{\textbf{Robustness}}.}
    \label{fig:mmperspective}
    \vspace{-3mm}
\end{figure}

\begin{abstract}
Understanding perspective is fundamental to human visual perception, yet the extent to which multimodal large language models (MLLMs) internalize perspective geometry remains unclear. We introduce MMPerspective, the first benchmark specifically designed to systematically evaluate MLLMs' understanding of perspective through 10 carefully crafted tasks across three complementary dimensions: Perspective Perception, Reasoning, and Robustness. Our benchmark comprises 2,711 real-world and synthetic image instances with 5,083 question-answer pairs that probe key capabilities, such as vanishing point perception and counting, perspective type reasoning, line relationship understanding in 3D space, invariance to perspective-preserving transformations, etc. Through a comprehensive evaluation of 43 state-of-the-art MLLMs, we uncover significant limitations: while models demonstrate competence on surface-level perceptual tasks, they struggle with compositional reasoning and maintaining spatial consistency under perturbations. Our analysis further reveals intriguing patterns between model architecture, scale, and perspective capabilities, highlighting both robustness bottlenecks and the benefits of chain-of-thought prompting. MMPerspective establishes a valuable testbed for diagnosing and advancing spatial understanding in vision-language systems. Resources are available at \url{https://yunlong10.github.io/MMPerspective/}
\end{abstract}

\section{Introduction}

\begin{quote}
\raggedright %
{\em Perspective is nothing more than a rational demonstration applied to the consideration of how objects in front of the eye transmit their image to it.}\\
\hfill-- {\scriptsize Leonardo da Vinci, \emph{The Notebooks of Leonardo da Vinci}~\citep{da2012notebooks}} %
\end{quote}

From the chalked strings of Renaissance artists to the calibrated optics of modern cameras, perspective has long served as a cornerstone for representing three-dimensional reality on two-dimensional surfaces~\citep{kemp1990science,neher2005perspective}.
Based on the geometry of the pinhole camera model, perspective projection enables humans to infer spatial structure, depth, and layout from flat images, a capability central to artistic creation, scientific visualization, and machine perception~\citep{hartley2003multiple,hecht2012optics}.
For instance, artists employ perspective to enhance realism, guide viewer attention, manipulate spatial illusion, and convey narrative depth~\citep{robertson2013draw,panofsky2020perspective}.
In scientific visualization, perspective projections are used to render complex 3D structures, such as molecular surfaces and anatomical forms~\citep{ware2019information}.
In computer vision, some methods based on the perspective principle have been developed to analyze, edit images, and fix distortions~\citep{criminisi2002bringing,carroll2010image,carroll2013warping}.
Therefore, perspective understanding plays a foundational role in visual cognition and spatial representation.
However, current research~\citep{bharadwaj2025recurrence,coudert2022semi,zhao2021deephough} is still primarily focused on using perspective principles to implement various applications, with relatively little research on the ability of intelligent systems themselves to understand perspective.
Although some studies have already attempted to enable models to locate vanishing points~\citep{bharadwaj2025recurrence}, detect key lines in space~\citep{coudert2022semi,zhao2021deephough}, etc., these models either rely on precise mathematical models or learn from specialized datasets, being hard to capture perspective-related semantics or apply their learned understanding of perspective to other more general tasks.

On the other hand, recent multimodal large language models (MLLMs) such as GPT-4o~\citep{achiam2023gpt4} and Gemini~\citep{reid2024gemini} have demonstrated powerful human-like visual perception and reasoning capabilities through large-scale training, but their ability to understand perspective has not yet been tested.
Given its foundational role in visual cognition and spatial representation, an important open question is: \textbf{Do MLLMs understand perspective?} These models have shown remarkable performance across a broad range of high-level vision-language tasks, including visual captioning~\citep{wang2023caption} and visual question answering~\citep{liu2024llava,achiam2023gpt4,reid2024gemini,chen2024internvl,Qwen2-VL}. However, existing benchmarks rarely evaluate their capacity for geometric reasoning. In particular, it remains unclear whether MLLMs can identify vanishing points, understand the convergence of parallel lines, reason about spatial relationships induced by perspective, or maintain consistent spatial interpretations across different viewpoints. These are fundamental aspects of human visual understanding and have been systematically studied in both art history and computational vision~\citep{robertson2013draw}, yet they are largely absent from current evaluation protocols~\citep{yu2024mmvetevaluatinglargemultimodal,liu2025mmbench,li2024mvbench, hua2024mmcomposition,wang2024lvbench,tang2024vidcomposition,tang2025video,tang2025videolmm_posttraining} for MLLMs.

To bridge this gap, we introduce \textbf{\NAME}, the first benchmark specifically designed to evaluate perspective understanding in MLLMs. As shown in \Cref{fig:mmperspective}, our benchmark comprises 10 tasks divided across three dimensions: \textbf{\textcolor{my_red}{Perspective Perception}}, \textbf{\textcolor{my_blue}{Perspective Reasoning}}, and \textbf{\textcolor{my_green}{Perspective Robustness}}. Perception tasks probe the ability to identify geometric cues such as vanishing points and critical lines. Reasoning tasks examine models' ability to interpret 3D structure, assess scene composition, and predict off-canvas geometry. Robustness task evaluates spatial consistency under appearance-preserving transformations, such as flipping and cropping.

Our benchmark comprises \textbf{2,711} image instances and \textbf{5,083} question-answer pairs, each framed as a multiple-choice question grounded in real-world imagery rich with architectural, urban, and indoor perspective cues, such as vanishing lines, orthogonal edges, and depth gradients. Tasks are organized to increase in difficulty across perceptual, reasoning, and robustness dimensions, requiring progressively deeper spatial abstraction.
We evaluate 43 state-of-the-art MLLMs, ranging from lightweight open-source models to proprietary systems like GPT-4o and Gemini. While many models perform competitively on surface-level perception tasks, they exhibit clear performance drops on reasoning and robustness tasks. For instance, models often fail to maintain consistent predictions under simple geometric-preserving edits, such as horizontal flipping or partial occlusion of key cues, revealing their limited internalization of spatial priors and geometric constraints.

\begin{figure}[!ht]
    \centering
    \includegraphics[width=0.95\linewidth]{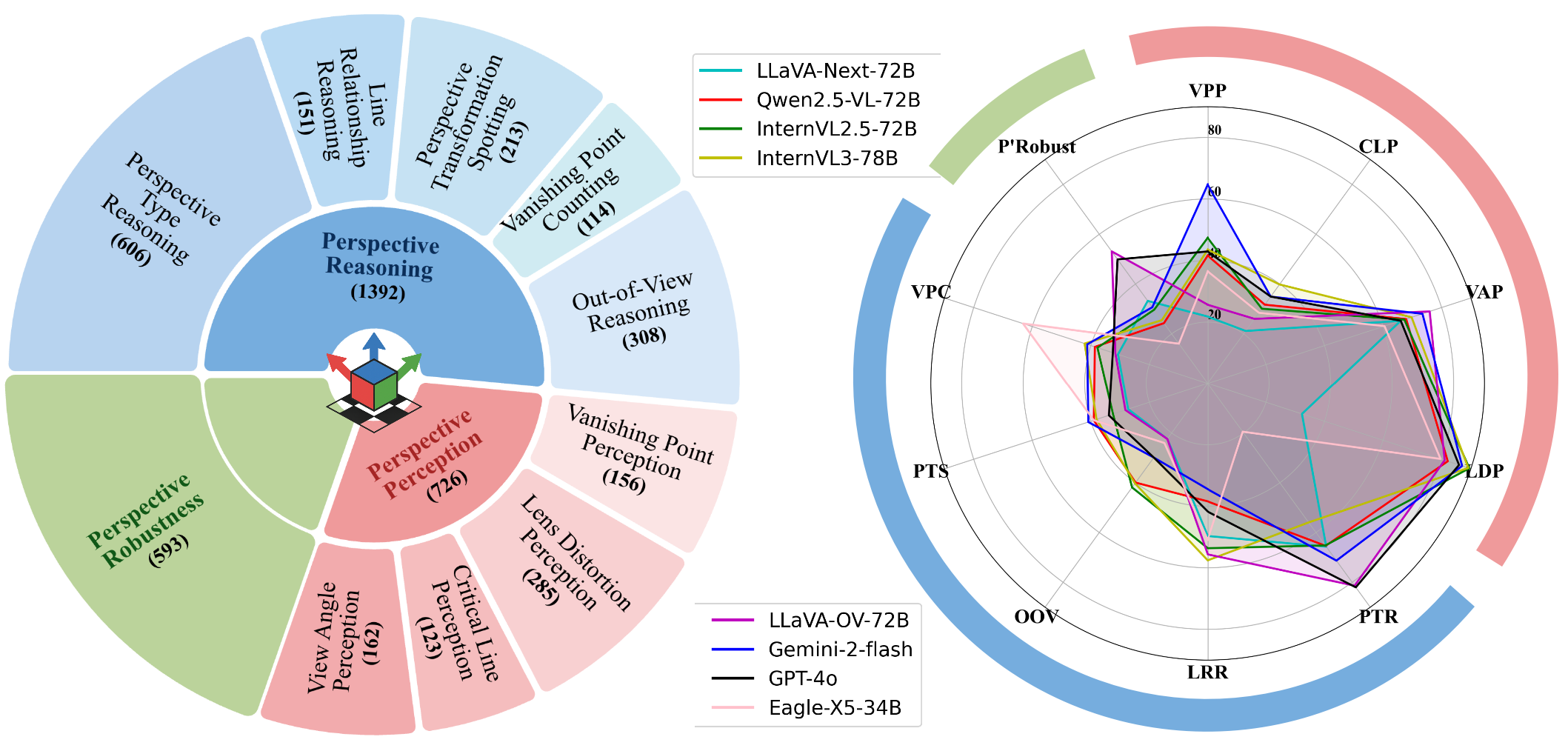}
    \vspace{-3mm}
    \caption{\textbf{Left}: \NAME benchmark consists of 2,711 instances and 5,083 QA pairs, hierarchically organized into 3 core categories (Perspective \textcolor{my_red}{\textbf{Perception}}, \textcolor{my_blue}{\textbf{Reasoning}}, and \textcolor{my_green}{\textbf{Robustness}}). \textbf{Right}: The accuracy of 8 representative MLLMs on 10 tasks of \NAME across the 3 categories.}
    \label{fig:radar}
\end{figure}

In short, our contributions are three-fold:
\begin{itemize}
\item We introduce \textbf{\NAME}, the first dedicated benchmark for evaluating perspective understanding in MLLMs, spanning 10 tasks across three dimensions, consisting of 2,711 instances and 5,083 QA pairs.
\item We conduct a comprehensive evaluation of 43 representative MLLMs and reveal key limitations in perspective perception, reasoning, and robustness.
\item We offer new insights into current model bottlenecks and provide guidance toward building geometry-aware, spatially grounded multimodal systems.
\end{itemize}

\section{\NAME}

\subsection{Preliminary}
\begin{wrapfigure}[13]{r}{0.55\linewidth}
    \vspace{-16mm}
    \centering
    \includegraphics[width=\linewidth]{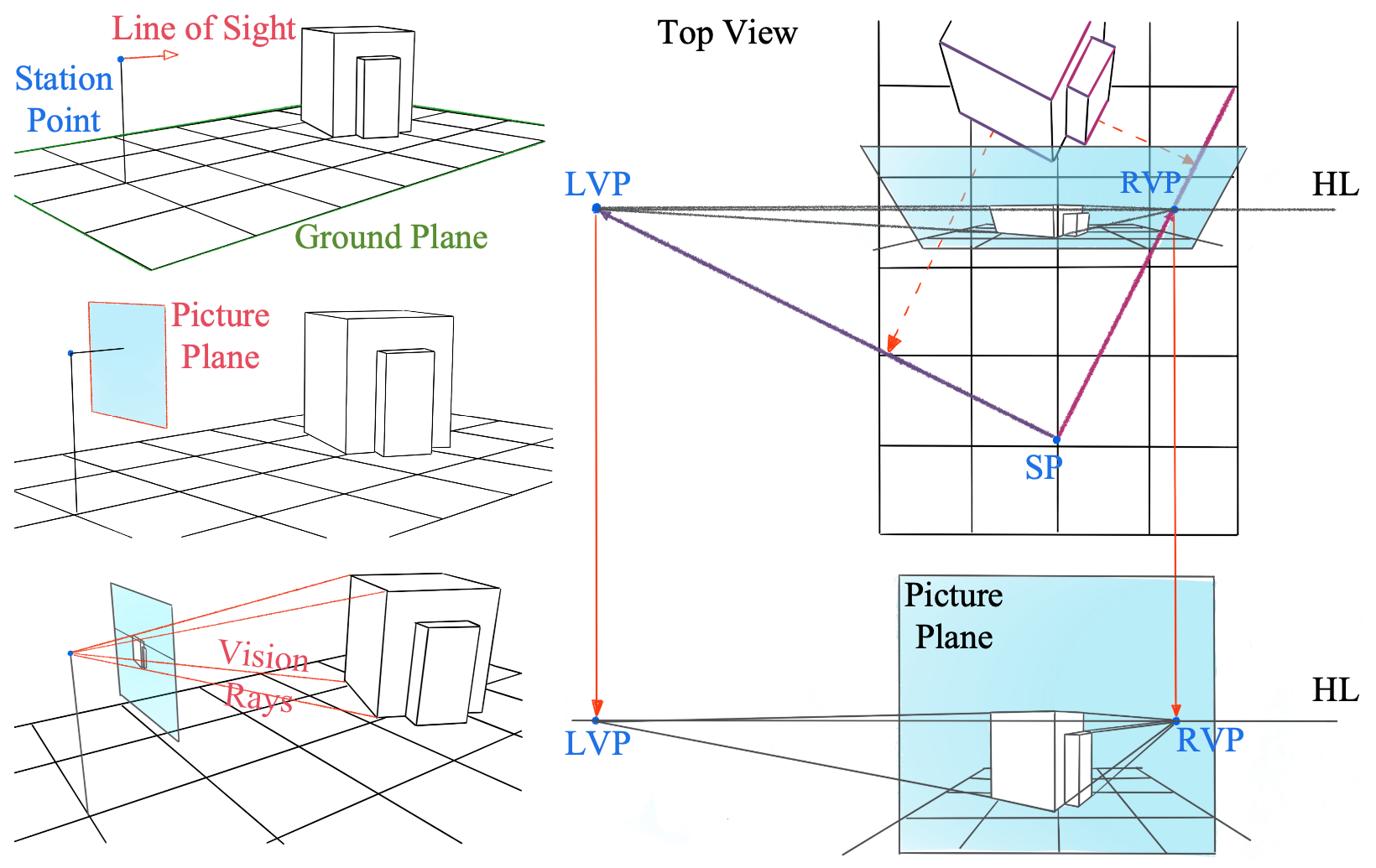}
    \vspace{-2mm}
    \caption{Perspective illustration with terminology. The figure is adapted from \citep{robertson2013draw}.}
    \label{fig:pre}
    \vspace{-4mm}
\end{wrapfigure}
Understanding the key elements of perspective geometry is essential for interpreting spatial relationships in 2D images. In this section, we introduce foundational terms used, following classical principles of linear perspective as described in drawing literature~\citep{robertson2013draw}.
As shown in the \Cref{fig:pre}, the \textbf{Ground Plane (GP)} is the surface upon which objects rest and from which vertical height is measured. The \textbf{Station Point (SP)} represents the viewer's position in space, typically aligned with the eye or camera origin. The \textbf{Line of Sight (LS)} defines the direction in which the observer is looking; when this is parallel to GP, vertical lines in the scene remain vertical in the image, as seen in one- or two-point perspectives. Tilting the LS results in three-point perspective, where verticals also converge.
The \textbf{Picture Plane (PP)} refers to an imaginary plane perpendicular to the LS where the visual projection occurs. It is often conceptualized as a transparent sheet placed between the observer and the scene, capturing the intersections of visual rays from the Station Point to the object. The \textbf{Vision Rays (VRs)} are the lines extending from the eye through each point on the object to the PP.
The \textbf{Horizon Line (HL)} corresponds to the viewer’s eye level and is the projection of the GP onto the PP. A \textbf{Vanishing Point (VP)} is the point at which a set of parallel lines appears to converge. In 1-point perspective, a single set of lines converges to one VP. In 2-point perspective, two sets of lines converge to separate VPs on the HL. In 3-point perspective, an additional VP is used for vertical convergence, located either above or below the HL, depending on whether the observer is looking up or down.


\subsection{Taxonomy}

The \NAME benchmark is designed to evaluate perspective understanding in MLLMs across three complementary and hierarchically structured dimensions: \textcolor{my_red}{\textbf{Perspective Perception}}, \textcolor{my_blue}{\textbf{Perspective Reasoning}}, and \textcolor{my_green}{\textbf{Perspective Robustness}}. These dimensions reflect a progression from low-level visual recognition to high-level spatial inference and consistency under image transformations.

\noindent\textbf{\textcolor{my_red}{Perspective Perception} (\textcolor{my_red}{P'Percep})} focuses on a model’s ability to detect and interpret explicit perspective-related cues directly visible in the image. It includes the following tasks:
\textbf{Vanishing Point Perception (\textcolor{my_red}{VPP})} evaluates whether a model can correctly locate a VP or determine its presence within a given region.
\textbf{Critical Line Perception (\textcolor{my_red}{CLP})} assesses the identification of the HL from a set of candidate lines, based on perspective convergence.
\textbf{Lens Distortion Perception (\textcolor{my_red}{LDP})} requires the model to distinguish regions in the image that are free from curved-line distortion.
\textbf{View Angle Perception (\textcolor{my_red}{VAP})} asks the model to infer the LS direction (e.g., upward, downward, or horizontal) using visible spatial cues.
All tasks in this category are grounded in localized, directly observable visual evidence and require minimal reasoning beyond geometric feature detection.

\noindent\textbf{\textcolor{my_blue}{Perspective Reasoning} (\textcolor{my_blue}{P'Reason)}}  tests whether the model can integrate multiple spatial cues and apply geometric reasoning to infer high-level relationships in the 3D structure of the scene. The tasks include:
\textbf{Perspective Type Reasoning (\textcolor{my_blue}{PTR})}, which involves classifying the underlying perspective structure of the image (e.g., 1-point, 2-point, 3-point, or non-linear).
\textbf{Line Relationship Reasoning (\textcolor{my_blue}{LRR})}, which asks the model to determine whether two lines in the 3D space are parallel, perpendicular, or intersecting.
\textbf{Perspective Transformation Spotting (\textcolor{my_blue}{PTS})}, which requires detecting changes in perspective type across paired images.
\textbf{Vanishing Point Counting (\textcolor{my_blue}{VPC})}, which involves estimating the number of identifiable VPs present in the scene.
\textbf{Out-of-View Reasoning (\textcolor{my_blue}{OVR})}, which challenges the model to infer the quadrant in which a VP lies when it is not explicitly shown in the image.
These tasks demand a combination of compositional reasoning, global geometric understanding, and spatial abstraction beyond direct visual perception.

\noindent\textbf{\textcolor{my_green}{Perspective Robustness} (\textcolor{my_green}{P'Robust})} assesses the model’s ability to produce consistent and geometry-aware predictions under controlled, appearance-preserving transformations of the input image. Each original image-question pair is augmented with perturbed versions through perspective-invariant operations such as cropping, flipping, and masking. While these transformations do not alter the scene’s underlying geometry, they may obscure or de-emphasize key visual cues.
A model is considered robust if it provides the same, correct answer across all such transformed variants. This consistency serves as a direct measure of its geometric grounding, separating genuine perspective understanding from brittle reliance on surface-level visual patterns.


\begin{figure}[!ht]
    \centering
    \includegraphics[width=0.95\linewidth]{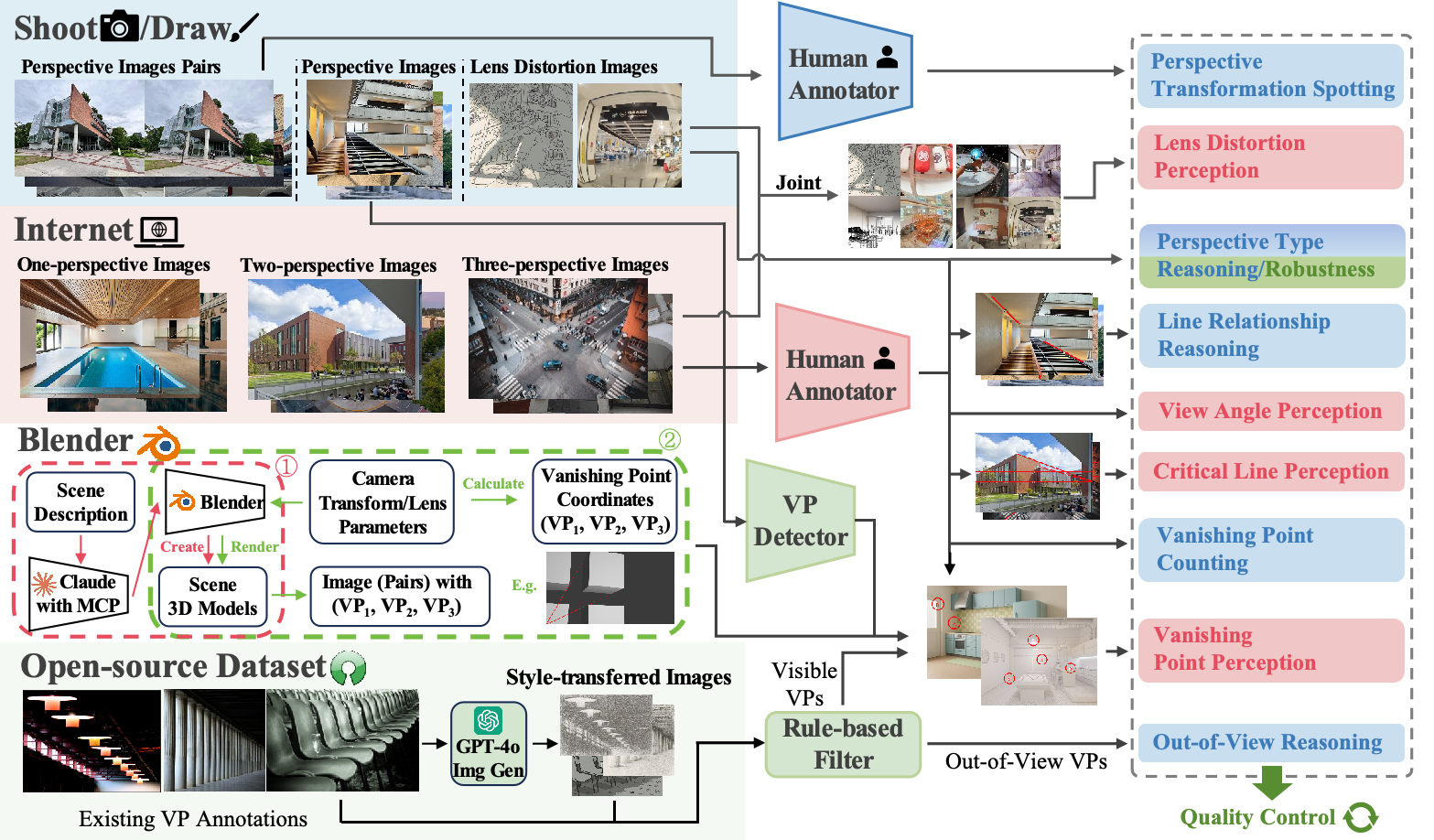}
    \caption{Data Curation Pipeline for \NAME.}
\end{figure}

\subsection{Data Curation} \label{sec:data_curation}
\noindent\textbf{Data Collection.}
To support the construction of these tasks, we curated a diverse set of perspective-rich images from multiple sources. Images are sourced from four streams. \textbf{First}, we collect unlabeled examples from the web, primarily architectural and indoor scenes with strong perspective cues. \textbf{Second}, we shoot real-world perspective images in life scenarios with both linear perspectives and curvilinear perspectives (fish-eye perspectives). For one scene, we shoot multiple images with different views to form perspective image pairs. \textbf{Third}, we incorporate data from the open-source RPVP datasets~\citep{bharadwaj2025recurrence}. In this dataset, perspective cues come from the recurrence pattern rather than lines at object edges. \textbf{Fourth}, we utilize Blender to create images with ground-truth VP coordinates. Specifically, we first employ Claude 3.7 Sonnet to create 3D models based on scene descriptions, empowered by Blender-MCP. For each scene, we render multiple images with different camera transform and lens parameters. From these parameters, we calculate the ground-truth VP coordinates for each image. We provide more details of this approach in Appendix.

\noindent\textbf{Annotation.}
We annotate each image with task-specific metadata using a hybrid pipeline. For \textcolor{my_blue}{\textbf{PTS}}, we manually annotate the perspective changes in the image pairs that we shoot. For \LDP, we combine fish-eye perspective images and regular linear perspective images randomly and record the corresponding option. For \PTR, \LRR, \VAP, \CLP, and \VPC, we use images collected from the web and manually annotate the right answers for the questions and hints on the images. For \VPP, we use both images from the web and Blender. The VP annotations of the former are manually created, while the latter are born with ground-truth VP coordinates. For \OVR, we use the annotation from the RPVP datasets~\citep{bharadwaj2025recurrence}.

\noindent\textbf{Quality Control.}
Quality assurance is carried out via a multi-stage review process. All automatically generated annotations are verified manually. For subjective tasks involving spatial reasoning, at least two annotators independently label each sample, with disagreements resolved through discussion and consensus. We exclude any examples where ambiguity could not be resolved, and the final benchmark comprises only unambiguous, perspective-defining scenes. We also manually check and filter all unsafe images we collect.

\subsection{Evaluation Metrics}

For all tasks in \PPercep~and \PReason~of \NAME, we use accuracy as the main evaluation metric, where each question has one correct answer. For \PRobust, we evaluate consistency under image perturbations and report two complementary metrics:

\paragraph{Binary \textbf{P'Robust} Score.}
Let $\mathcal{S}$ be the set of robustness seed items. For each seed $(I_s, q, a^*)\in\mathcal{S}$ we consider the set of images
$V_s=\{I_s\}\cup\{I_1,\dots,I_{n_s}\}$ which includes the original image and all its perturbed variants.
Binary robustness requires perfect consistency across all images in $V_s$:
\begin{equation}
\text{Binary-}\mathrm{Robust}_{\mathcal{M}}
= \frac{1}{|\mathcal{S}|}\sum_{(I_s,q,a^*)\in\mathcal{S}}
\mathbbm{1}\!\left[\,\bigwedge_{I\in V_s}\mathcal{M}(I,q)=a^*\,\right].
\label{eq:binary-robust}
\end{equation}

\paragraph{Graded \textbf{P'Robust} Score.}
To capture partial consistency, we additionally compute a graded score that averages the fraction of correctly answered images within each set $V_s$:
\begin{equation}
\text{Graded-}\mathrm{Robust}_{\mathcal{M}}
= \frac{1}{|\mathcal{S}|}\sum_{(I_s,q,a^*)\in\mathcal{S}}
\left(\frac{1}{|V_s|}\sum_{I\in V_s}\mathbbm{1}\!\left[\mathcal{M}(I,q)=a^*\right]\right).
\label{eq:graded-robust}
\end{equation}
For example, if a model answers $4$ out of $5$ images in $V_s$ correctly, its per-set graded score is $0.8$, while its binary score for that set would be $0$.

\section{Experiments}

\subsection{Experiment Setup} \label{sec:exp_setup}
We select 20 representative models, including both open-source and proprietary models, covering a broad spectrum of model scales and architecture types. These include GPT-4o~\citep{openai2024gpt4ocard}, Gemini-2~\citep{deepmind_gemini_flash}, LLaVA-OV~\citep{li2024llavaonevisioneasyvisualtask}, LLaVA-Next~\citep{liu2024llavanext}, InternVL2~\citep{chen2024internvl}, InternVL2.5~\citep{chen2024internvl}, InternVL3~\citep{zhu2025internvl3}, Qwen2-VL~\citep{wang2024qwen2}, Qwen2.5-VL~\citep{bai2025qwen25vltechnicalreport}, and Eagle-X~\citep{shi2024eagle}.
To ensure fairness and eliminate potential positional bias, we have already randomly shuffled the answer choices for all questions during the dataset creation process.
To ensure consistency, all open-source models under 14B are evaluated using a single NVIDIA A6000 48GB GPU. Models larger than 14B and up to 70B are evaluated using a single NVIDIA H100 80GB GPU. Larger models (>70B) are run on multiple NVIDIA A100 80G GPUs (at least 4). Proprietary models are executed via APIs. Each model is evaluated under the same test conditions, with identical multiple-choice question formats across all tasks. To ensure deterministic and fully reproducible results for all our experiments, we employed a greedy decoding strategy for all open-source models. For proprietary models accessed via API, we also used their deterministic decoding modes where available. This approach eliminates randomness from the decoding process, ensuring that a model's output for any given sample is consistent across multiple runs. 

\subsection{Main Results}

\begin{table}[!ht]
\caption{\textbf{Performance of MLLMs on MMPerspective.} Models are grouped by size and ranked by overall accuracy. Best scores in each group are bolded.}
\label{tab:mm_perspective_performance}
\vspace{2mm}
\centering
\resizebox{\textwidth}{!}{
\begin{tabular}{l|cccc|ccccc|ccc|cc}
\toprule
& \multicolumn{4}{c|}{\textbf{Perspective Perception}} & \multicolumn{5}{c|}{\textbf{Perspective Reasoning}} & \multicolumn{3}{c|}{\textbf{\textcolor{my_red}{P'Percep} \& \textcolor{my_blue}{P'Reason}}} & \multicolumn{2}{c}{\textbf{\textcolor{my_green}{Robustness}}} \\
\cmidrule(lr){2-5} \cmidrule(lr){6-10} \cmidrule(lr){11-13} \cmidrule(lr){14-15}
\textbf{Model} & \textbf{\textcolor{my_red}{VPP}} & \textbf{\textcolor{my_red}{CLP}} & \textbf{\textcolor{my_red}{VAP}} & \textbf{\textcolor{my_red}{LDP}} & \textbf{\textcolor{my_blue}{PTR}} & \textbf{\textcolor{my_blue}{LRR}} & \textbf{\textcolor{my_blue}{OVR}} & \textbf{\textcolor{my_blue}{PTS}} & \textbf{\textcolor{my_blue}{VPC}} & \textbf{P Acc} & \textbf{R Acc} & \textbf{Overall} & \textbf{Graded} & \textbf{Binary} \\
\midrule
\rowcolor[HTML]{e9edf6}
\multicolumn{15}{c}{\textit{MLLMs:} < 7B} \\
\midrule
\textbf{InternVL2.5-2B} & \textbf{47.4} & 22.8 & 13.0 & \textbf{65.3} & \textbf{62.2} & 31.8 & 16.6 & 30.0 & \textbf{50.0} & 37.1 & \textbf{38.1} & \textbf{37.7} & \textbf{59.1} & \textbf{46.5} \\
\textbf{Qwen2.5-VL-3B} & 27.6 & 22.8 & 56.8 & 55.1 & 32.3 & 32.5 & 15.9 & \textbf{39.4} & 44.7 & 40.6 & 33.0 & 36.3 & 22.2 & 6.4 \\
\textbf{InternVL2.5-4B} & 32.1 & 26.0 & \textbf{59.3} & 64.2 & 28.2 & 30.5 & 10.7 & 37.1 & 36.8 & \textbf{45.4} & 28.7 & 36.1 & 25.0 & 20.6 \\
\textbf{InternVL3-2B} & 22.4 & \textbf{28.5} & 50.0 & 44.6 & 43.1 & 31.1 & \textbf{34.4} & 25.4 & 43.0 & 36.4 & 35.4 & 35.8 & 39.0 & 23.9 \\
\textbf{InternVL2-4B} & 26.9 & 12.2 & 54.3 & 60.4 & 18.0 & \textbf{40.4} & 18.8 & 24.4 & 45.6 & 38.4 & 29.4 & 33.4 & 14.5 & 7.9 \\
\textbf{Qwen2-VL-2B} & 12.2 & 19.5 & 49.4 & 35.8 & 23.3 & 24.5 & 28.9 & 32.9 & 47.4 & 29.2 & 31.4 & 30.4 & 18.0 & 4.7 \\
\textbf{InternVL3-1B} & 19.9 & 13.0 & 53.7 & 20.7 & 16.3 & 8.6 & 23.7 & 21.6 & 47.4 & 26.8 & 23.5 & 25.0 & 16.1 & 13.8 \\
\textbf{InternVL2-1B} & 20.5 & 20.3 & 15.4 & 24.2 & 24.1 & 11.3 & 24.0 & 22.1 & 44.7 & 20.1 & 25.2 & 23.0 & 18.2 & 6.7 \\
\textbf{LLaVA-OV-1B} & 13.5 & 14.6 & 35.8 & 24.2 & 15.2 & 19.2 & 19.5 & 22.1 & 40.4 & 22.0 & 23.3 & 22.7 & 13.0 & 7.8 \\
\textbf{InternVL2-2B} & 26.9 & 26.0 & 3.1 & 36.8 & 18.8 & 12.6 & 23.1 & 21.1 & 34.2 & 23.2 & 22.0 & 22.5 & 19.3 & 12.3 \\
\textbf{InternVL2.5-1B} & 14.7 & 23.6 & 0.6 & 33.0 & 20.1 & 11.3 & 13.3 & 34.7 & 45.6 & 18.0 & 25.0 & 21.9 & 19.0 & 18.2 \\
\midrule
\rowcolor[HTML]{e9edf6}
\multicolumn{15}{c}{\textit{MLLMs:} 7B - 9B} \\
\midrule
\textbf{InternVL2.5-8B} & 38.5 & 17.9 & 53.1 & 75.4 & 40.8 & 48.3 & 34.7 & 24.9 & 67.5 & 46.2 & 43.3 & \textbf{44.6} & 38.7 & 22.3 \\
\textbf{Qwen2.5-VL-7B} & 35.3 & 29.3 & \textbf{70.4} & 73.7 & 42.4 & 44.4 & 32.1 & 28.6 & 44.7 & 52.1 & 38.5 & 44.5 & 33.2 & 15.3 \\
\textbf{Qwen2-VL-7B} & 34.6 & 25.2 & 63.0 & 64.2 & 57.1 & 49.0 & 27.3 & 31.0 & 46.5 & 46.7 & 42.2 & 44.2 & 46.9 & 25.5 \\
\textbf{InternVL3-9B} & 37.2 & \textbf{33.3} & 63.0 & 77.5 & 30.7 & \textbf{53.0} & 27.9 & 23.9 & 43.9 & 52.8 & 35.9 & 43.4 & 19.2 & 7.3 \\
\textbf{InternVL3-8B} & \textbf{42.3} & 27.6 & 67.9 & \textbf{81.8} & 38.1 & 46.4 & 20.8 & 23.9 & 32.5 & \textbf{54.9} & 32.3 & 42.4 & 29.1 & 15.9 \\
\textbf{LLaVA-OV-7B} & 34.0 & \textbf{33.3} & 51.2 & 57.9 & 44.9 & \textbf{53.0} & 19.8 & \textbf{35.2} & 49.1 & 44.1 & 40.4 & 42.0 & 36.1 & 15.9 \\
\textbf{Eagle-X4-8B} & 39.1 & 17.1 & 46.9 & 47.7 & \textbf{65.3} & 37.1 & 18.2 & 32.9 & \textbf{68.4} & 37.7 & \textbf{44.4} & 41.4 & \textbf{60.7} & \textbf{55.3} \\
\textbf{InternVL2-8B} & 33.3 & 19.5 & 59.3 & 73.3 & 27.1 & 36.4 & \textbf{42.5} & 22.1 & 48.2 & 46.4 & 35.3 & 40.2 & 19.9 & 7.9 \\
\textbf{LLaVA-Next-m-7B} & 35.9 & 21.1 & 35.2 & 50.5 & 17.7 & 37.7 & 15.6 & 27.2 & 46.5 & 35.7 & 28.9 & 31.9 & 17.9 & 16.4 \\
\textbf{Eagle-X5-7B} & 25.0 & 26.0 & 24.7 & 34.7 & 22.1 & 46.4 & 15.6 & 20.7 & 42.1 & 27.6 & 29.4 & 28.6 & 18.4 & 15.9 \\
\textbf{LLaVA-Next-v-7B} & 16.7 & 20.3 & 40.7 & 39.6 & 16.3 & 44.4 & 19.8 & 16.4 & 7.0 & 29.3 & 20.8 & 24.6 & 16.7 & 16.4 \\
\midrule
\rowcolor[HTML]{e9edf6}
\multicolumn{15}{c}{\textit{MLLMs:} 10B - 30B} \\
\midrule
\textbf{InternVL2.5-26B} & 41.7 & \textbf{35.0} & 55.6 & \textbf{81.8} & 65.5 & \textbf{46.4} & 43.5 & \textbf{34.3} & 46.5 & \textbf{53.5}& \textbf{47.2} & \textbf{50.0} & 52.9 & 33.7 \\
\textbf{InternVL3-14B} & 39.1 & 26.0 & \textbf{73.5} & 73.3 & 36.5 & 34.4 & \textbf{54.5} & 28.2 & 54.4 & 53.0 & 41.6 & 46.7 & 27.3 & 13.5 \\
\textbf{InternVL2-26B} & 28.2 & \textbf{35.0} & 61.1 & 74.0 & 50.7 & 41.7 & 28.9 & 28.6 & 43.0 & 49.6 & 38.6 & 43.5 & 44.1 & 26.5 \\
\textbf{Eagle-X4-13B} & \textbf{42.3} & 26.8 & 41.4 & 44.6 & 65.8 & 20.5 & 28.2 & 31.0 & \textbf{57.9} & 38.8 & 40.7 & 39.8 & \textbf{60.7} & \textbf{53.8} \\
\textbf{LLaVA-Next-13B} & 7.7 & 17.1 & 54.3 & 34.7 & \textbf{66.7} & 24.5 & 13.0 & 26.8 & 43.9 & 28.5 & 35.0 & 32.1 & 59.7 & 51.1 \\
\midrule
\rowcolor[HTML]{e9edf6}
\multicolumn{15}{c}{\textit{MLLMs:} 30B - 70B} \\
\midrule
\textbf{InternVL2.5-38B} & \textbf{46.8} & \textbf{36.6} & 67.9 & 89.5 & 58.4 & 51.7 & 38.3 & \textbf{44.1} & 44.7 & 60.2 & \textbf{47.5} & \textbf{53.1} & 41.6 & 19.1 \\
\textbf{InternVL3-38B} & 45.5 & 35.0 & \textbf{71.0} & \textbf{90.9} & 37.3 & 43.0 & \textbf{56.8} & 37.6 & 43.0 & \textbf{60.6} & 43.5 & 51.1 & 23.9 & 9.1 \\
\textbf{Qwen2.5-VL-32B} & 35.9 & 22.8 & 68.5 & 73.7 & \textbf{62.0} & 37.7 & 33.8 & 35.2 & 45.6 & 50.2 & 42.9 & 46.1 & \textbf{48.8} & \textbf{25.5} \\
\textbf{Eagle-X5-34B} & 36.5 & 28.5 & 60.5 & 79.6 & 19.5 & 51.0 & 24.0 & 39.0 & \textbf{63.2} & 51.3 & 39.3 & 44.6 & 18.7 & 16.0 \\
\textbf{InternVL2-40B} & 26.3 & 22.0 & 66.0 & 76.1 & 43.2 & \textbf{55.0} & 27.3 & 25.8 & 47.4 & 47.6 & 39.7 & 43.2 & 29.5 & 12.6 \\
\midrule
\rowcolor[HTML]{e9edf6}
\multicolumn{15}{c}{\textit{MLLMs:} > 70B} \\
\midrule
\textbf{InternVL3-78B} & 43.6 & \textbf{39.8} & 69.8 & 89.1 & 55.9 & \textbf{57.6} & 40.3 & 38.0 & \textbf{42.1} & \textbf{60.6} & 46.8 & \textbf{52.9} & 43.6 & 25.5 \\
\textbf{InternVL2.5-72B} & \textbf{47.4} & 30.1 & 67.3 & \textbf{89.5} & 65.2 & 53.6 & \textbf{41.9} & 32.4 & 37.7 & 58.6 & 46.2 & 51.7 & 56.3 & 29.7 \\
\textbf{Qwen2.5-VL-72B} & 41.7 & 31.7 & 67.9 & 82.1 & 65.3 & 38.4 & 39.9 & \textbf{39.0} & 38.6 & 55.8 & 44.3 & 49.4 & 49.9 & 24.3 \\
\textbf{Qwen2-VL-72B} & 34.6 & 18.7 & 70.4 & 82.5 & 68.8 & 52.3 & 38.6 & 35.2 & \textbf{42.1} & 51.5 & \textbf{47.4} & 49.2 & 51.3 & 25.0 \\
\textbf{LLaVA-OV-72B} & 25.6 & 26.0 & \textbf{75.9} & 81.1 & \textbf{81.4} & 55.6 & 22.4 & 28.2 & 31.6 & 52.2 & 43.8 & 47.5 & \textbf{71.8} & \textbf{53.1} \\
\textbf{LLaVA-Next-72B} & 21.8 & 21.1 & 66.0 & 32.3 & 65.7 & 49.7 & 22.4 & 27.2 & 30.7 & 35.3 & 39.1 & 37.4 & 55.6 & 33.2 \\
\textbf{InternVL2-72B} & 26.9 & 18.7 & 57.4 & 56.8 & 56.1 & 47.0 & 24.7 & 24.4 & 7.9 & 40.0 & 32.0 & 35.6 & 43.9 & 22.9 \\
\midrule
\rowcolor[HTML]{e9edf6}
\multicolumn{15}{c}{\textit{MLLMs:} Proprietary} \\
\midrule
\textbf{Gemini-2-flash (CoT)} & \textbf{69.2} & \textbf{49.6} & 72.8 & 87.4 & 78.7 & 32.5 & \textbf{40.9} & 39.9 & 43.9 & \textbf{69.8} & \textbf{47.2} & \textbf{57.2} & 50.5 & 24.8 \\
GPT-4o (CoT) & 45.5 & 46.3 & 70.4 & \textbf{88.8} & 81.4 & \textbf{47.0} & 34.4 & 37.6 & 34.2 & 62.7 & 46.9 & 54.0 & 69.4 & \textbf{49.9}  \\
\textbf{Gemini-2-flash} & 64.7 & 35.0 & \textbf{73.5} & 87.0 & 71.3 & 34.4 & 29.9 & \textbf{40.8} & 41.2 & 65.0 & 43.5 & 53.1 & 56.8 & 30.7 \\
\textbf{GPT-4o} & 42.9 & 35.0 & 66.0 & 86.0 & \textbf{82.0} & 41.7 & 29.9 & 33.8 & 32.5 & 57.5 & 44.0 & 50.0 & \textbf{71.9} & \textbf{49.9} \\
\textbf{Gemini-1.5-flash (CoT)} & 30.1 & 28.5 & 66.7 & 79.3 & 51.0 & 39.7 & 20.1 & 31.5 & 35.1 & 51.1 & 35.5 & 42.4 & 37.8 & 11.6 \\
\textbf{GPT-4o-mini} & 35.3 & 24.4 & 43.2 & 71.6 & 43.1 & 29.8 & 14.6 & 31.0 & \textbf{45.6} & 43.6 & 32.8 & 37.6 & 28.7 & 10.8 \\
\textbf{Gemini-1.5-flash} & 26.9 & 25.2 & 59.3 & 70.5 & 26.4 & 27.8 & 18.2 & 26.8 & 22.8 & 45.5 & 24.4 & 33.8 & 20.6 & 10.6 \\
\bottomrule
\end{tabular}
}
\end{table}

Table~\ref{tab:mm_perspective_performance} presents the performances of various MLLMs on our \textsc{MMPerspective} benchmark. In general, larger models tend to perform better, with GPT-4o and Gemini-2-flash achieving the highest overall accuracy (57.7\% and 57.6\%, respectively).

\noindent
\textbf{Perspective Perception.}
For \VPP, Gemini-2-flash (CoT) achieves the highest accuracy (69.8\%), while many smaller models struggle with this fundamental task. In \CLP, all models perform poorly, with even GPT-4o (CoT) only reaching 46.3\%, indicating a general limitation in detecting HLs. Most larger models exceed 60\% on \VAP, with InternVL3-14B leading at 73.5\%. For \LDP, InternVL3-38B demonstrates the strongest performance (90.9\%), surpassing even proprietary models.

\noindent
\textbf{Perspective Reasoning.}
In \PTR, GPT-4o achieves the highest score (82.0\%), with LLaVA-OV-72B close behind (81.4\%). \LRR~shows less correlation with model size, with InternVL3-78B leading at 57.6\%. For \OVR, InternVL3-38B significantly outperforms all others (56.8\%), suggesting unique architectural advantages. In \VPC, the Eagle-X4 family demonstrates superior performance (68.4\% for 8B), indicating specialized capabilities for identifying multiple VPs.

\noindent
\textbf{Perspective Robustness.}
\PRobust~scores reveal surprising patterns, with Eagle-X4-8B achieving great performance (55.3\%) despite modest size. LLaVA-OV-72B (53.1\%) and Eagle-X4-13B (53.8\%) also present strong robustness. Notably, many large models with high accuracy perform poorly on robustness, with InternVL3-38B showing excellent perception (67.2\%) but poor robustness (9.1\%).

\begin{figure}[!ht]
    \centering
    \includegraphics[width=0.8\linewidth]{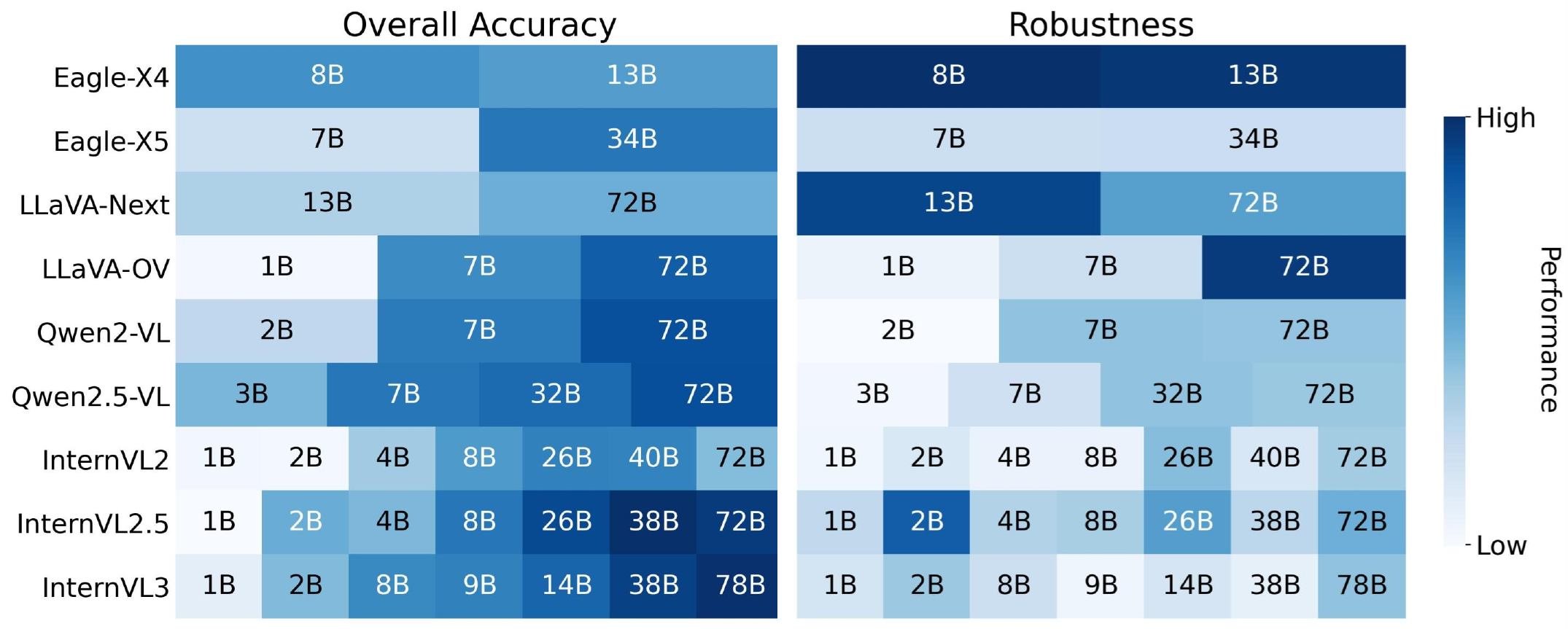}
    \caption{Heatmaps illustrating the relationship between model size and performance, measured by P\&R Overall Accuracy and Robustness. Darker colors indicate higher performance. Each line represents a model family, with sizes increasing from left to right.}
    \label{fig:heatmaps}
\end{figure}

\begin{figure}[!ht]
    \centering
    \includegraphics[width=\linewidth]{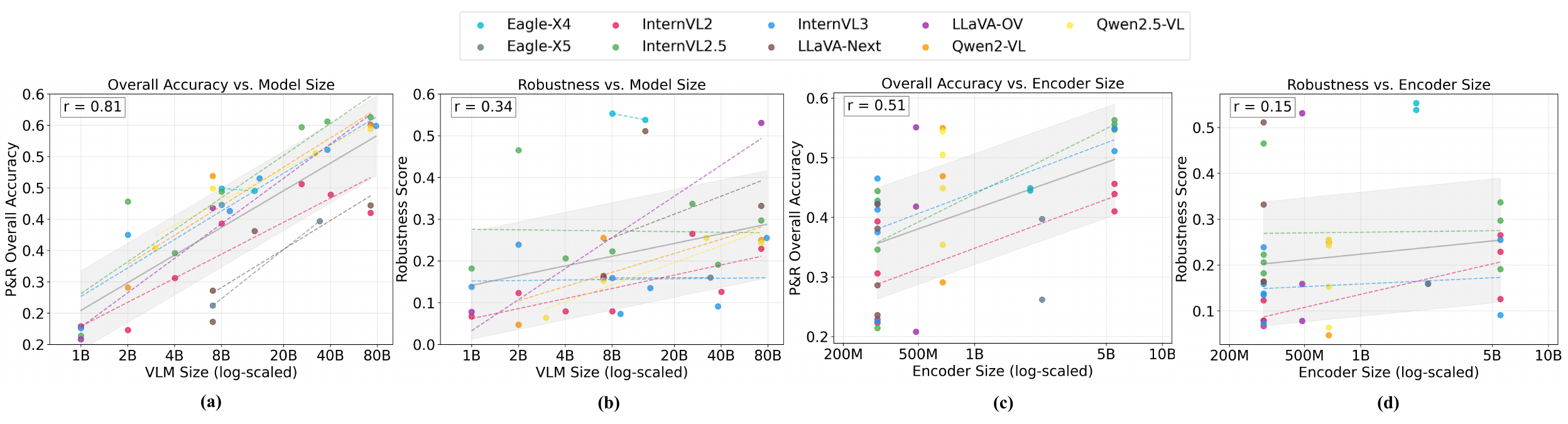}
    \caption{Correlation analysis between performance and size across MLLM families: (a) Overall accuracy vs. model size ($r = 0.81$), (b) Robustness vs. model size ($r = 0.34$), (c) Overall accuracy vs. encoder size ($r = 0.51$), (d) Robustness vs. encoder size ($r = 0.15$). Total model scaling strongly impacts perspective understanding, while vision encoder size has a limited influence on robustness.}
    \label{fig:scatter_vision_2}
\end{figure}

\subsection{Further Findings}

\noindent
\begin{tcolorbox}[colframe=black,
arc=2pt,
boxsep=-0.35em,
left=8pt,right=8pt,
]
\paragraph{\textbf{Finding 1.}} Our analysis reveals that perspective understanding scales strongly with total model size but only weakly with vision encoder size, with robustness showing particularly limited correlation to encoder scaling.
\end{tcolorbox}

Our analysis of model scaling reveals important insights into how different architectural components influence perspective understanding capabilities in MLLMs. In Fig.~\ref{fig:heatmaps}, there is a clear progression of performance within model families as model size increases, with deeper blue coloration indicating higher accuracy and robustness for larger variants. The scatter plots in Fig.~\ref{fig:scatter_vision_2} quantify these relationships more precisely, demonstrating a strong positive correlation between model size and perspective understanding accuracy ($r = 0.81$), while robustness shows a weaker correlation ($r = 0.34$).

This disparity suggests that while general perspective understanding capabilities scale reliably with language model size, robustness to perspective-preserving transformations follows a different pattern. For instance, models like Eagle-X4 achieve high perspective robustness even at moderate sizes (8B and 13B), suggesting their architecture may have inherent advantages for maintaining consistent geometric interpretations across image variations.

When examining vision encoder scaling specifically (Fig.~\ref{fig:scatter_vision_2}c-d), we observe a moderate correlation with overall perspective accuracy ($r = 0.51$) but a notably weak correlation with perspective robustness ($r = 0.15$). This suggests that vision encoders play a more limited role in ensuring consistent geometric interpretations across transformations than in enabling basic perspective understanding. The data indicates that while increasing vision encoder capacity may help models better recognize perspective features initially, it does not necessarily translate to more stable geometric interpretations when those features are partially obscured or repositioned.

The limited range of encoder sizes currently employed across model families (mostly 300-500M parameters) makes it difficult to draw definitive conclusions about vision encoder scaling laws for perspective understanding. This represents a gap in our understanding of how to optimally design MLLMs for spatial reasoning tasks that require both accurate perspective perception and consistent geometric interpretations under varying conditions. 


\noindent
\begin{tcolorbox}[colframe=black,
arc=2pt,
boxsep=-0.35em,
left=8pt,right=8pt,
]
\paragraph{\textbf{Finding 2.}} Chain-of-thought (CoT) prompting modestly improves model performance and robustness on perspective-related tasks by encouraging stepwise deduction.
\end{tcolorbox}

\begin{table*}[t]
\caption{\textbf{Chain of Thought (CoT) prompting improves MLLM performance on perspective tasks.} Accuracy changes due to CoT prompting across perception and reasoning tasks.}
\label{tab:cot_improvements}
\centering
\resizebox{\textwidth}{!}{
\begin{tabular}{l|cccc|ccccc|ccc|c}
\toprule
 & \multicolumn{4}{c|}{\textbf{Perspective Perception}} & \multicolumn{5}{c|}{\textbf{Perspective Reasoning}} & \multicolumn{3}{c|}{\textbf{\PPercep~\& \PReason}} & \PRobust \\
\midrule
 & \VPP & \CLP & \VAP & \LDP & \PTR & \LRR & \OVR & \PTS & \VPC & \textbf{P Acc} & \textbf{R Acc} & \textbf{Overall} & \textbf{Binary} \\
\midrule
\textbf{GPT-4o} & +2.56 & +11.38 & +4.32 & +2.81 & -0.66 & +5.30 & +4.55 & +3.76 & +1.75 & +5.27 & +2.94 & +3.97 & +0.00 \\
\textbf{Gemini-1.5-flash} & +3.21 & +3.25 & +7.41 & +8.77 & +24.59 & +11.92 & +1.95 & +4.69 & +12.28 & +5.66 & +11.09 & +8.67 & +4.72 \\
\textbf{Gemini-2-flash} & +4.49 & +14.63 & -0.62 & +0.35 & +7.43 & -1.99 & +11.04 & -0.94 & +2.63 & +4.71 & +3.63 & +4.11 & +15.18 \\
\midrule
\textbf{Average $\Delta$} & +3.42 & +9.76 & +3.70 & +3.98 & +10.45 & +5.08 & +5.84 & +2.50 & +5.56 & +5.21 & +5.89 & +5.59 & +6.63 \\
\bottomrule
\end{tabular}
}

\vspace{-5mm}
\end{table*}

As shown in Table~\ref{tab:cot_improvements}, CoT prompting leads to consistent performance gains across nearly all perspective-related tasks. All three evaluated models, GPT-4o, Gemini-1.5-flash, and Gemini-2-flash, experience improvements in both perception and reasoning sub-tasks when CoT is applied. Notably, no single sub-task exhibits degradation in performance for more than one model, suggesting that CoT prompting is broadly beneficial and rarely harmful within this domain.

The overall accuracy and robustness metrics also trend upward with CoT, reinforcing its value not only in structured reasoning but also in enhancing the model's resilience to perspective-related perturbations. For instance, the average gain in P\&R Overall Accuracy is +5.59\%, and in Robustness is +6.63\%, indicating that step-by-step reasoning contributes to more confident and stable outputs.

While the benefits are widespread, a few failures still emerge. In Appendix, we analyze three representative failure cases to better understand CoT's limitations. These include GPT-4o on Perspective Type Reasoning, and Gemini-2-flash on Line Relationship and Perspective Transformation Spotting. 


Overall, our findings suggest that while CoT prompting is not a silver bullet, it provides meaningful and reliable improvements in most perspective tasks. This points toward the promise of integrating structured reasoning strategies with visual understanding, especially for tasks where spatial interpretation and viewpoint deduction are required.

\begin{figure}[!ht]
    \centering
    \includegraphics[width=1\linewidth]{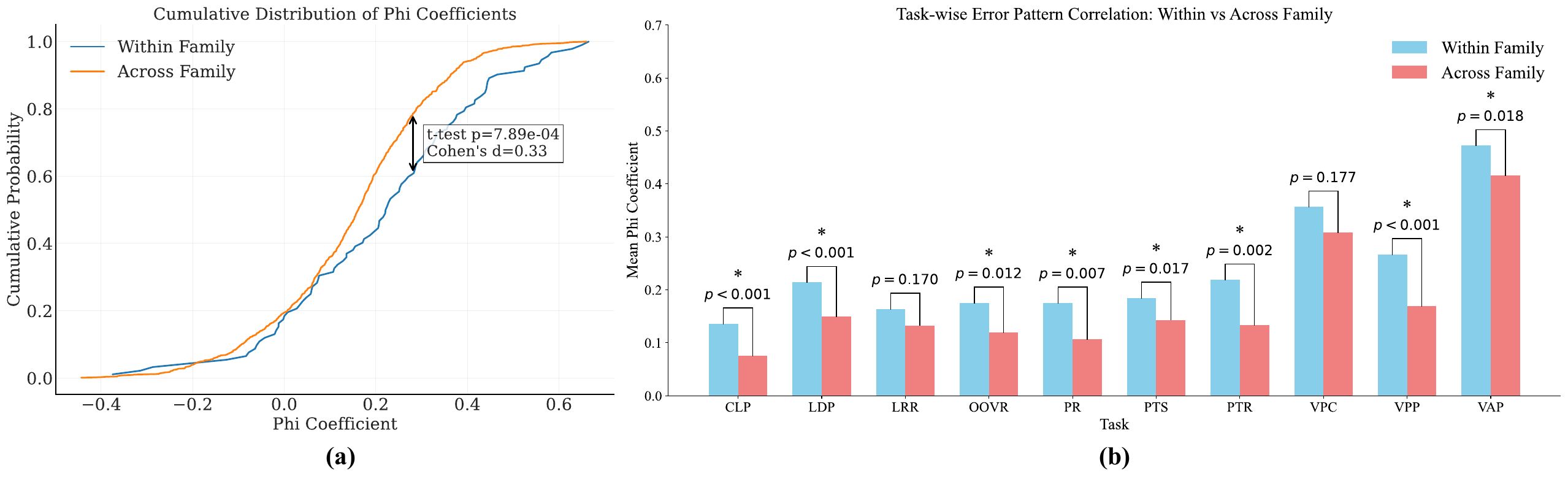}
    \caption{Error pattern analysis across model families: (a) Cumulative distribution of phi coefficients shows significantly higher correlations within families than across families (Cohen's $d=0.33$, $p<0.001$). (b) Task-wise breakdown reveals perception tasks (\VAP, \CLP) exhibit the strongest family-specific patterns, while reasoning tasks (\VPC, \LRR) show weaker family effects.}
    \label{fig:error_patterns}
\end{figure}

\noindent
\begin{tcolorbox}[colframe=black,
arc=2pt,
boxsep=-0.35em,
left=8pt,right=8pt,
]
\paragraph{\textbf{Finding 3.}} Error pattern analysis reveals that while architectural/training choices strongly influence perspective perception biases, some spatial reasoning challenges present consistent difficulties across all model families.
\end{tcolorbox}

Error correlations reveal that model architecture/training strongly influences perspective understanding failure modes. Fig.~\ref{fig:error_patterns}a demonstrates models from the same family exhibit significantly more similar error patterns than models from different families (Cohen's $d=0.33$, $p<0.001$), indicating architectural biases systematically affect perspective interpretation. The task-wise analysis in Fig.~\ref{fig:error_patterns}b reveals this family effect varies markedly across the perspective hierarchy: low-level perception tasks show the strongest architecture/training-specific biases, with \VAP~and \CLP~exhibiting the largest within-family versus across-family differences ($p<0.001$). Notably, some tasks maintain relatively high correlation coefficients even in cross-family comparisons, particularly for \VAP~(0.41) and \VPC~(0.31). This suggests certain perspective challenges present universal difficulties that transcend architectural/training differences, especially tasks requiring complex spatial judgment (\VAP) or precise counting of geometric features (\VPC). In contrast, tasks like \CLP~show much larger gaps between within-family and cross-family correlations, indicating these capabilities are more sensitive to architectural design or training choices. These patterns reveal that while architecture significantly shapes perspective understanding biases, some fundamental spatial reasoning challenges remain consistently difficult across model designs.



\section{Related Work}

\subsection{Perspective Understanding}
Perspective is a cornerstone of visual realism, dictating how objects in a 2D image are perceived as three-dimensional. The theory of perspective can be traced back to Renaissance art, where principles such as VPs and HLs were formalized~\citep{elkins1994poetics, haley2018perspective}. In computer graphics and vision, perspective projection ensures that parallel lines in the real world converge at a VP on the image plane~\citep{hartley2003multipleview}. Multiple VPs, depending on the orientation of objects, define 1-point, 2-point, or 3-point perspectives.
Efficient and accurate VP detection has been a critical area of research, facilitating tasks like scene reconstruction~\citep{lee2009geometry, hedau2009recovering} and camera calibration~\citep{zhang2000flexiblecalibration}. Techniques such as the Hough Transform~\citep{duda1972hough} and its extensions~\citep{candes2006houghvariants} enable robust line detection, while Gaussian sphere mapping~\citep{Barnard1983GaussianSphere} provides a framework for detecting intersections representing VPs. Classical methods often detect VPs through line segment intersections~\citep{quan1989determining, lutton1994contribution}, followed by clustering approaches~\citep{mclean1995vanishing} or specialized voting schemes~\citep{gamba1996vanishing}. Recent works leverage deep learning, with methods like NeurVPS~\citep{zhou2019neurvps} that employ conic convolution operators and the Deep Hough Transform~\citep{lin2022deep} to improve accuracy in VP detection across diverse datasets.

\subsection{Evaluation Benchmarks for MLLMs}
With the rapid advancement of MLLMs~\citep{fei2024multimodal}, numerous benchmarks have emerged to systematically evaluate diverse capabilities~\citep{li2025benchmark}. These benchmarks generally assess two dimensions: text-centric evaluations measuring commonsense knowledge and reasoning (MMMU~\citep{yue2024mmmu}, NaturalBench~\citep{li2024naturalbench}), and vision-centric assessments focusing on perception and robustness (MMBench~\citep{liu2024mmbench}, MME~\citep{fu2024mmecomprehensiveevaluationbenchmark}, Grit~\citep{gupta2022grit}). Specialized visual tasks are evaluated through benchmarks for spatial relationship comprehension (SEED-Bench~\citep{li2023seed}, MM-Vet~\citep{yu2023mm}), chart understanding (MMSTAR~\citep{chen2024we}, MuirBench~\citep{wang2024muirbench}), visual grounding (Flickr30k~\citep{plummer2015flickr30k}, TRIG~\citep{li2025visualtextgroundingmultimodal}), and hallucination detection (POPE~\citep{li2023evaluating}, HallusionBench~\citep{guan2024hallusionbench}). Common evaluation approaches include image captioning~\citep{lin2014microsoft, OnoeDocci2024}, Visual Question Answering~\citep{antol2015vqa, marino2019ok, Mathew2020DocVQAAD}, and visual reasoning~\citep{johnson2017clevr, Suhr2017nlvr,hua2025finematch}. However, while certain benchmarks incorporate deeper assessments of perspective understanding remains limited~\citep{thrush2022winoground,hua2024mmcomposition}.
\section{Conclusion}
In this work, we introduce \NAME, the first benchmark to systematically evaluate perspective understanding in MLLMs. Through 10 tasks across perception, reasoning, and robustness, we reveal that while current models demonstrate basic geometric awareness, they fall short in compositional reasoning and maintaining consistency under perspective-preserving transformations. Our large-scale evaluation of 43 models uncovers clear performance trends and architectural limitations, pointing to the need for stronger spatial priors and geometry-aware design. \NAME provides a foundation for diagnosing perspective-related weaknesses and guiding the development of more spatially grounded vision-language systems.

\bibliographystyle{plainnat}
\bibliography{reference}

\newpage
\section*{NeurIPS Paper Checklist}
\begin{enumerate}

\item {\bf Claims}
    \item[] Question: Do the main claims made in the abstract and introduction accurately reflect the paper's contributions and scope?
    \item[] Answer: \answerYes{} 
    \item[] Justification: We confirm it.

\item {\bf Limitations}
    \item[] Question: Does the paper discuss the limitations of the work performed by the authors?
    \item[] Answer: \answerYes{} 
    \item[] Justification: We discuss the limitation in the appendix.

\item {\bf Theory assumptions and proofs}
    \item[] Question: For each theoretical result, does the paper provide the full set of assumptions and a complete (and correct) proof?
    \item[] Answer: \answerNA{} 
    \item[] Justification: The paper does not include theoretical results. 

    \item {\bf Experimental result reproducibility}
    \item[] Question: Does the paper fully disclose all the information needed to reproduce the main experimental results of the paper to the extent that it affects the main claims and/or conclusions of the paper (regardless of whether the code and data are provided or not)?
    \item[] Answer: \answerYes{} 
    \item[] Justification: We provide our experiment setting in \Cref{sec:exp_setup}.

\item {\bf Open access to data and code}
    \item[] Question: Does the paper provide open access to the data and code, with sufficient instructions to faithfully reproduce the main experimental results, as described in supplemental material?
    \item[] Answer: \answerYes{} 
    \item[] Justification: We provide open-source data.

\item {\bf Experimental setting/details}
    \item[] Question: Does the paper specify all the training and test details (e.g., data splits, hyperparameters, how they were chosen, type of optimizer, etc.) necessary to understand the results?
    \item[] Answer: \answerYes{} 
    \item[] Justification: We provide our experiment setting in \Cref{sec:exp_setup}.

\item {\bf Experiment statistical significance}
    \item[] Question: Does the paper report error bars suitably and correctly defined or other appropriate information about the statistical significance of the experiments?
    \item[] Answer: \answerNo{} 
    \item[] Justification: Error bars are not reported because it would be computationally expensive.

\item {\bf Experiments compute resources}
    \item[] Question: For each experiment, does the paper provide sufficient information on the computer resources (type of compute workers, memory, time of execution) needed to reproduce the experiments?
    \item[] Answer: \answerYes{} 
    \item[] Justification: We provide the information in \Cref{sec:exp_setup}.
    
\item {\bf Code of ethics}
    \item[] Question: Does the research conducted in the paper conform, in every respect, with the NeurIPS Code of Ethics \url{https://neurips.cc/public/EthicsGuidelines}?
    \item[] Answer: \answerYes{} 
    \item[] Justification: We confirm it.

\item {\bf Broader impacts}
    \item[] Question: Does the paper discuss both potential positive societal impacts and negative societal impacts of the work performed?
    \item[] Answer: \answerNA{} 
    \item[] Justification: We think that there is no societal impact of the work performed.
    
\item {\bf Safeguards}
    \item[] Question: Does the paper describe safeguards that have been put in place for responsible release of data or models that have a high risk for misuse (e.g., pretrained language models, image generators, or scraped datasets)?
    \item[] Answer: \answerYes{} 
    \item[] Justification: We describe it in \Cref{sec:data_curation}.

\item {\bf Licenses for existing assets}
    \item[] Question: Are the creators or original owners of assets (e.g., code, data, models), used in the paper, properly credited and are the license and terms of use explicitly mentioned and properly respected?
    \item[] Answer: \answerYes{} 
    \item[] Justification: We confirm it.

\item {\bf New assets}
    \item[] Question: Are new assets introduced in the paper well documented and is the documentation provided alongside the assets?
    \item[] Answer: \answerYes{} 
    \item[] Justification: We confirm it.

\item {\bf Crowdsourcing and research with human subjects}
    \item[] Question: For crowdsourcing experiments and research with human subjects, does the paper include the full text of instructions given to participants and screenshots, if applicable, as well as details about compensation (if any)? 
    \item[] Answer: \answerNA{} 
    \item[] Justification: This work does not involve crowdsourcing.

\item {\bf Institutional review board (IRB) approvals or equivalent for research with human subjects}
    \item[] Question: Does the paper describe potential risks incurred by study participants, whether such risks were disclosed to the subjects, and whether Institutional Review Board (IRB) approvals (or an equivalent approval/review based on the requirements of your country or institution) were obtained?
    \item[] Answer: \answerNA{} 
    \item[] Justification: This work does not involve crowdsourcing.

\item {\bf Declaration of LLM usage}
    \item[] Question: Does the paper describe the usage of LLMs if it is an important, original, or non-standard component of the core methods in this research? Note that if the LLM is used only for writing, editing, or formatting purposes and does not impact the core methodology, scientific rigorousness, or originality of the research, declaration is not required.
    \item[] Answer: \answerNA{} 
    \item[] Justification: LLMs were only used for editing (e.g., grammar, spelling), data processing/filtering, and facilitating/running experiments. They were not part of the core methodology or used in any important, original, or non-standard way.

\end{enumerate}

\newpage
\appendix
\section*{Appendix}

\section{Task Definitions}
\begin{figure}
    \centering
    \includegraphics[width=1\linewidth]{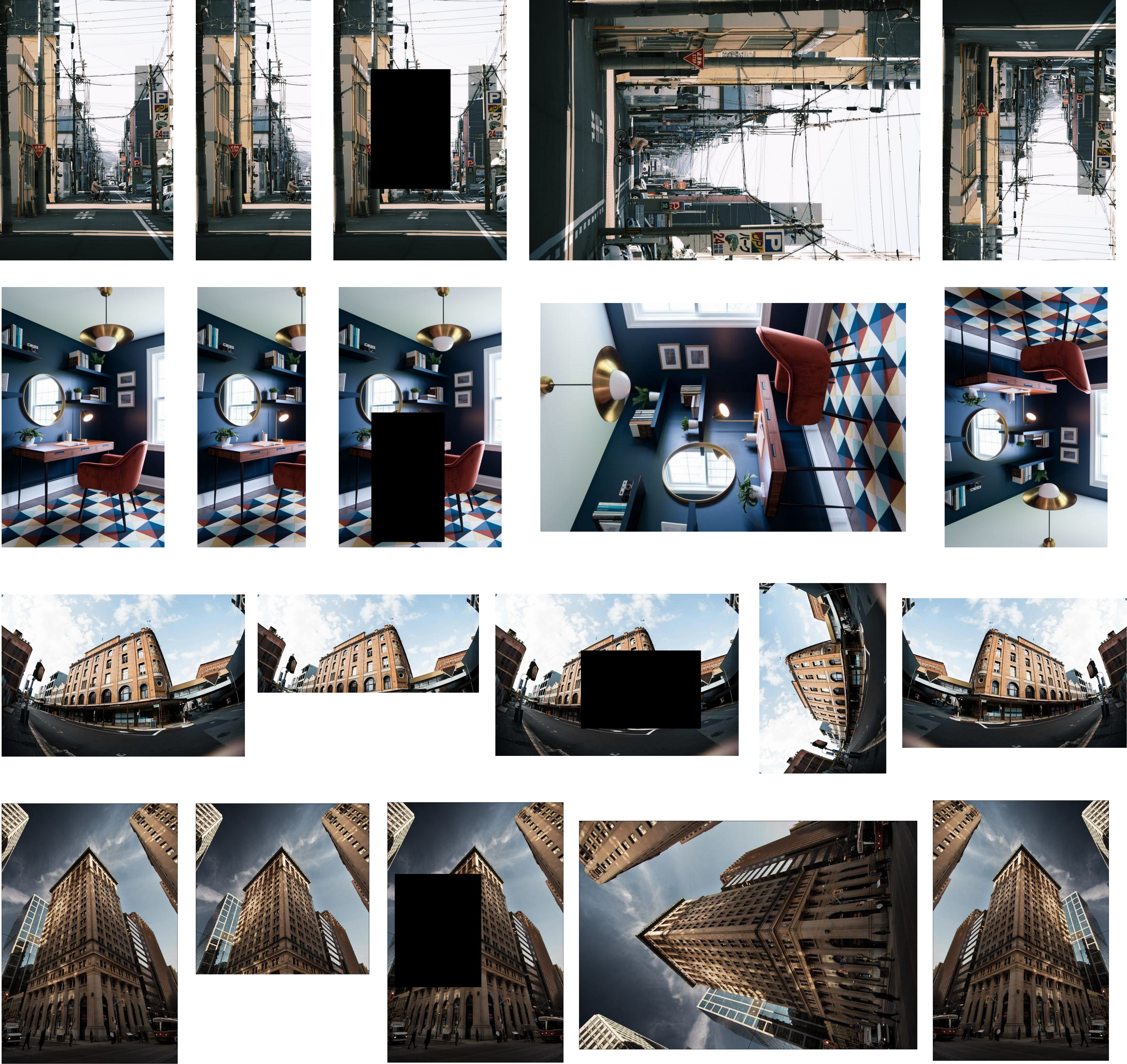}
    \caption{Examples of Perspective-Invariant Image Operations for Robustness Evaluation.}
    \label{fig:rob}
\end{figure}

\Cref{tab:task_description} outlines the and reasoning tasks included in the \NAME benchmark.
Sample cases and representative questions are included to illustrate the task format and input style.
We also show examples of perspective-invariant image operations for robustness evaluation in \Cref{fig:rob}, including cropping, masking, flipping, and rotation.
\begin{table}[!ht]
\centering
\caption{Task and question definition in \NAME.}
\label{tab:task_description}
\resizebox{\textwidth}{!}{
\begin{tabular}{@{}c|p{0.16\textwidth}cp{0.08\textwidth}p{0.2\linewidth}p{0.35\linewidth}}
\toprule
& \textbf{Task} & \textbf{\#} & \textbf{Sample Case} & \textbf{Description} & \textbf{Sample Questions} \\ 
\midrule
\multirow{3}{*}{\raisebox{-3cm}{\rotatebox[origin=c]{90}{Perspective Perception}}} 
& Vanishing Point Perception (\VPP) & 156 & Figure~\ref{fig:VPP} & Identify the region that contains the vanishing point in the image. & Where is the vanishing point in this image? \\ 
\cmidrule{2-6}
 & Critical Line Perception (\CLP) & 123 & Figure~\ref{fig:CLP} & Determine which of the highlighted lines is the horizon line. & Which line highlighted in the image is the Horizon Line? \\ 
\cmidrule{2-6}
 & View Angle Perception (\VAP) & 162 & Figure~\ref{fig:VAP} & Infer the camera’s line of sight direction from spatial cues. & What direction is the Line of Sight in this image? \\ 
\cmidrule{2-6}
 & Lens Distortion Perception (\LDP) & 285 & Figure~\ref{fig:LDP} & Identify the region without curved-line distortion in the image. & Which region shows no curved-line distortion? \\ 
\\ 
\midrule
\multirow{7}{*}{\raisebox{-4.5cm}{\rotatebox[origin=c]{90}{Perspective Reasoning}}} 
& Perspective Type Reasoning (\PTR) & 606 & Figure~\ref{fig:PTR} & Classify the perspective type used in the image (e.g., one-point, two-point). & What is the perspective type of this image? \\ 
\cmidrule{2-6}
 & Line Relationship Reasoning (\LRR) & 151 & Figure~\ref{fig:LRR} & Determine the spatial relationship between two lines in 3D (e.g., parallel, perpendicular). & What is the relationship between these two highlighted lines in the 3D space? \\ 
\cmidrule{2-6}
 & Perspective Transformation Spotting (\PTS) & 213 & Figure~\ref{fig:PTS} & Identify the change in perspective type between two images. & What changes occur from the left image to the right image? \\ 
\cmidrule{2-6}
 & Vanishing Point Counting (\VPC) & 114 & Figure~\ref{fig:VPC} & Count the number of vanishing points present in the image. & How many vanishing points can you identify within the image? \\ 
\cmidrule{2-6}
 & Out-of-View Reasoning (\OVR) & 308 & Figure~\ref{fig:OVR} & Infer the quadrant location of an unseen vanishing point based on scene geometry. & In which quadrant might the vanishing point be located? \\ 
\bottomrule
\end{tabular}
}
\end{table}

\begin{figure}
    \centering
    \includegraphics[width=1\linewidth]{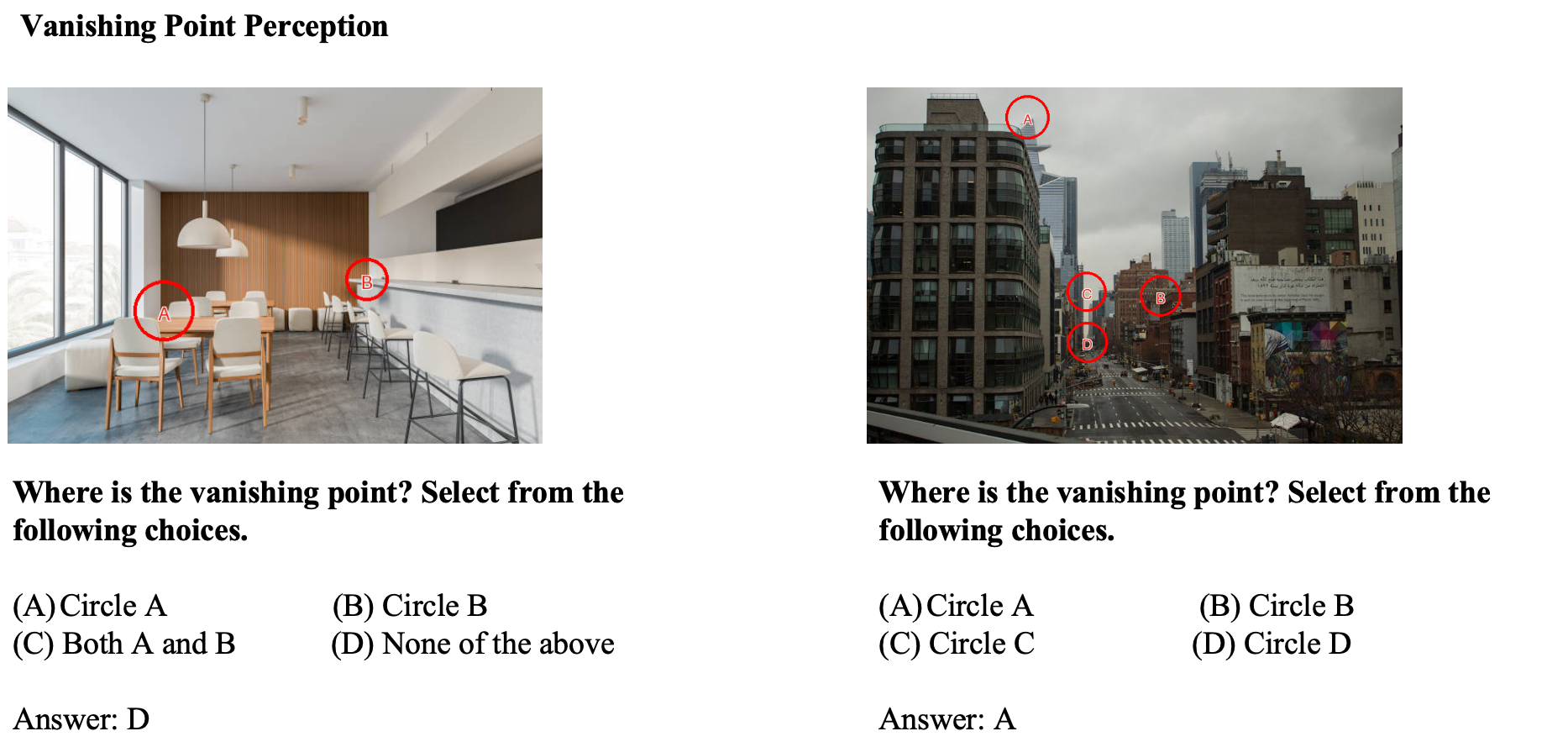}
    \caption{Examples of Vanishing Point Perception.}
    \label{fig:VPP}
\end{figure}
\begin{figure}
    \centering
    \includegraphics[width=1\linewidth]{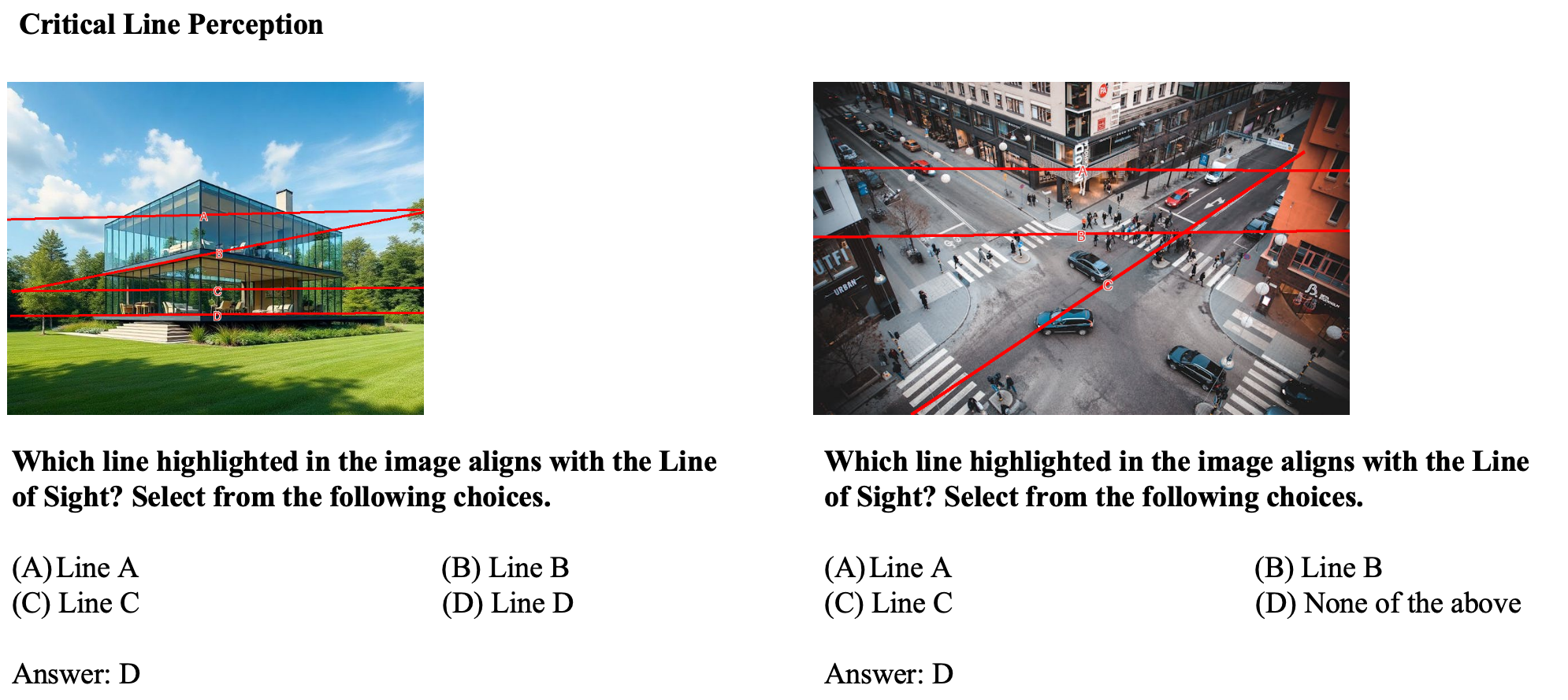}
    \caption{Examples of Critical Line Perception.}
    \label{fig:CLP}
\end{figure}
\begin{figure}
    \centering
    \includegraphics[width=1\linewidth]{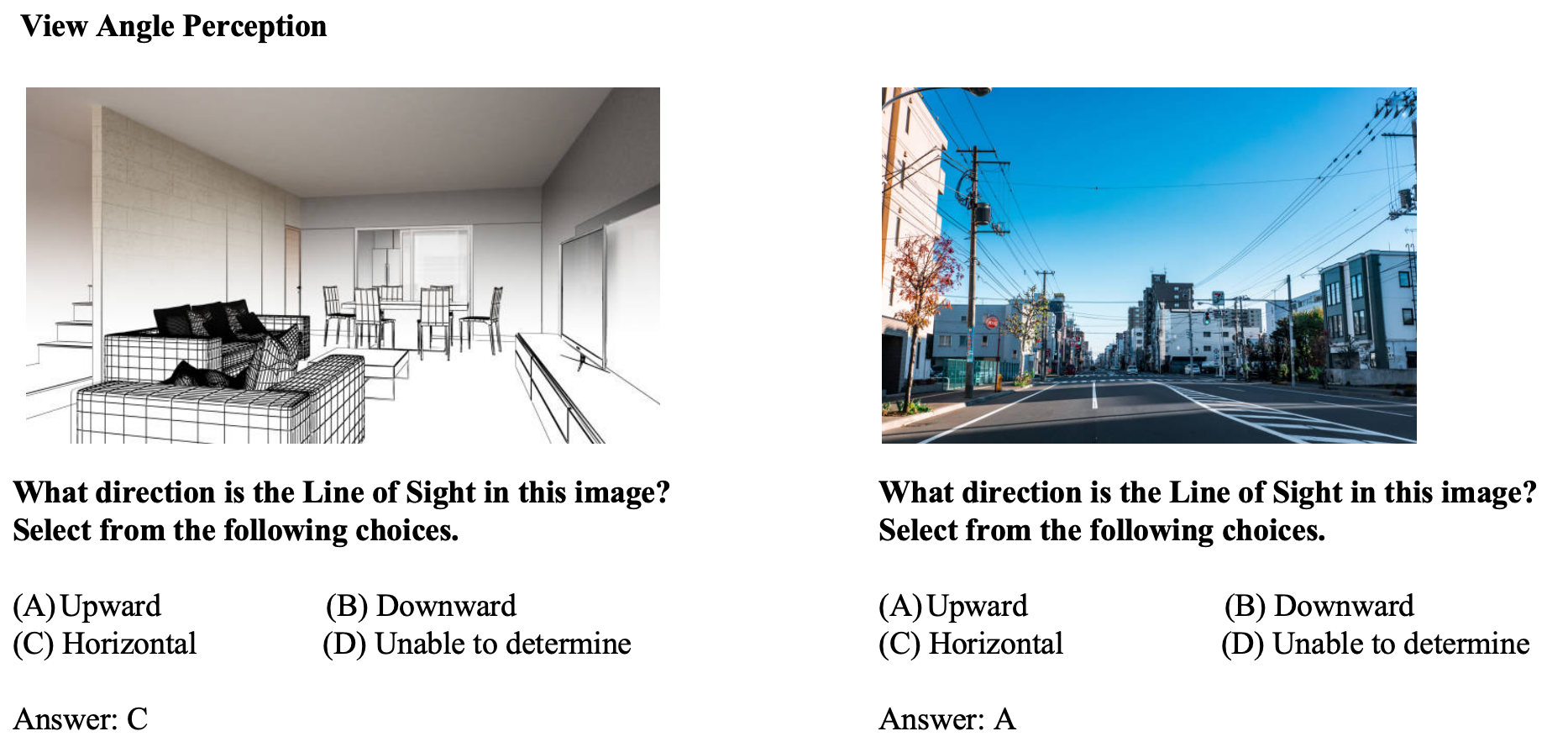}
    \caption{Examples of View Angle Perception.}
    \label{fig:VAP}
\end{figure}
\begin{figure}
    \centering
    \includegraphics[width=1\linewidth]{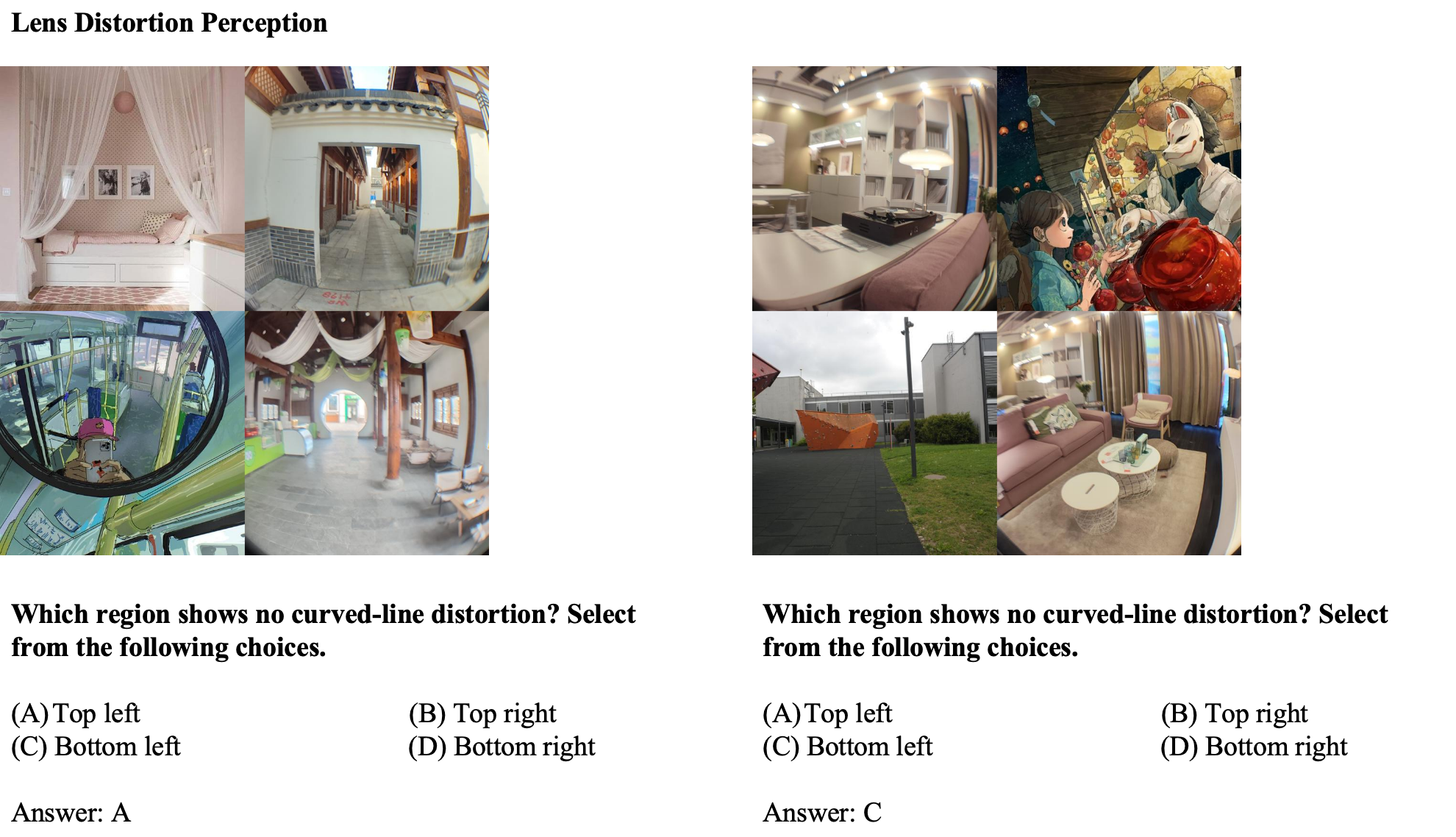}
    \caption{Examples of Line Relationship Reasoning.}
    \label{fig:LDP}
\end{figure}
\begin{figure}
    \centering
    \includegraphics[width=1\linewidth]{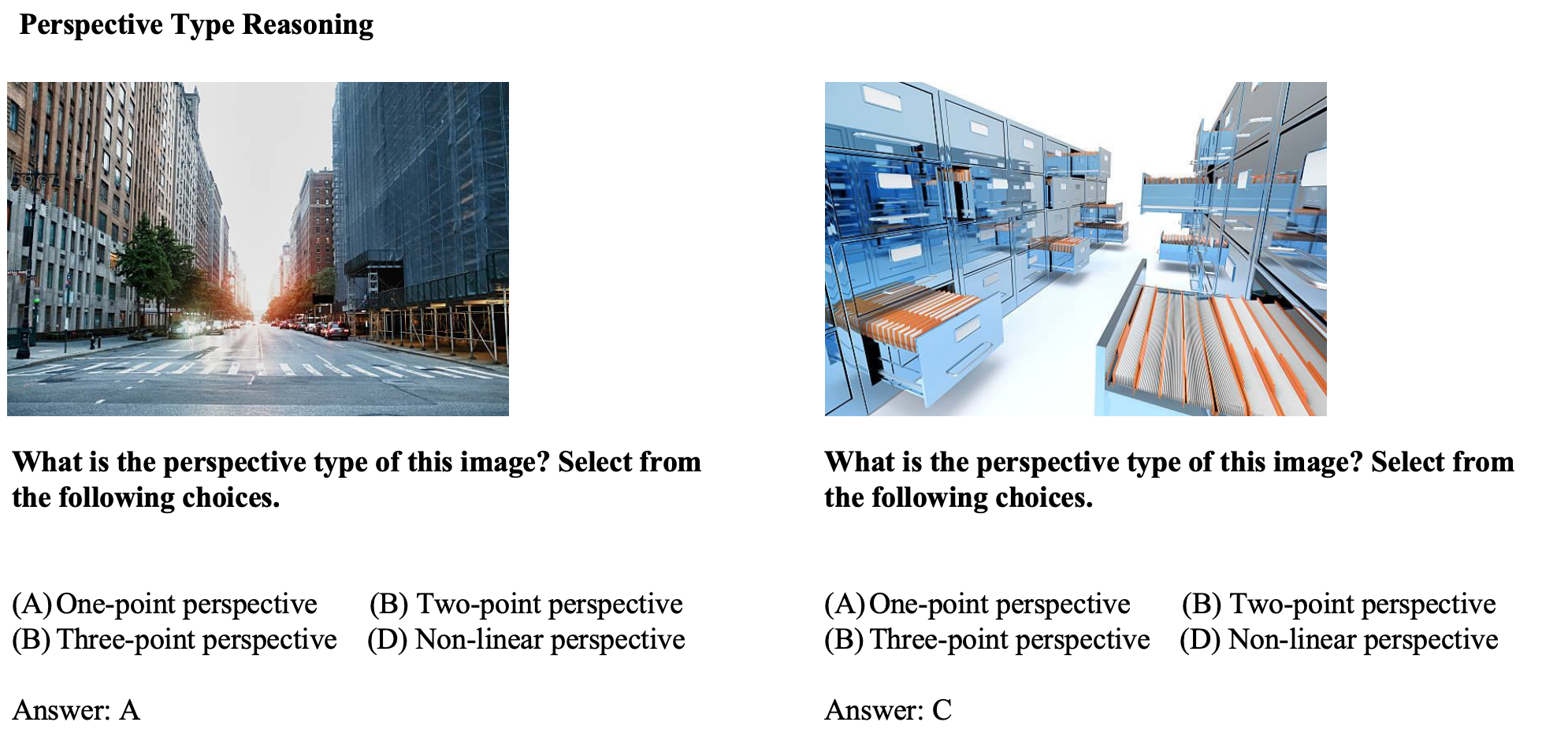}
    \caption{Examples of Perspective Type Reasoning.}
    \label{fig:PTR}
\end{figure}
\begin{figure}
    \centering
    \includegraphics[width=1\linewidth]{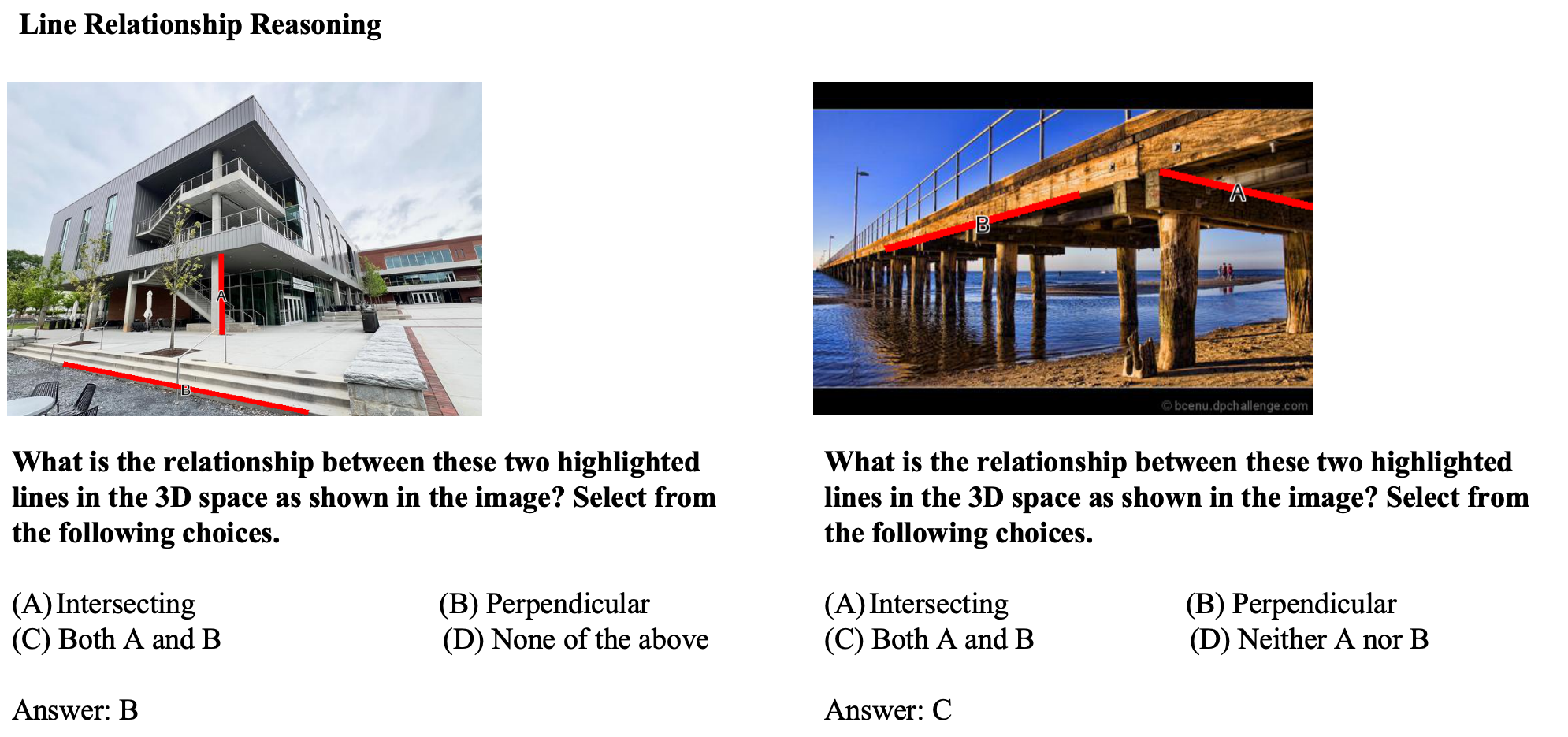}
    \caption{Examples of Line Relationship Reasoning.}
    \label{fig:LRR}
\end{figure}
\begin{figure}
    \centering
    \includegraphics[width=1\linewidth]{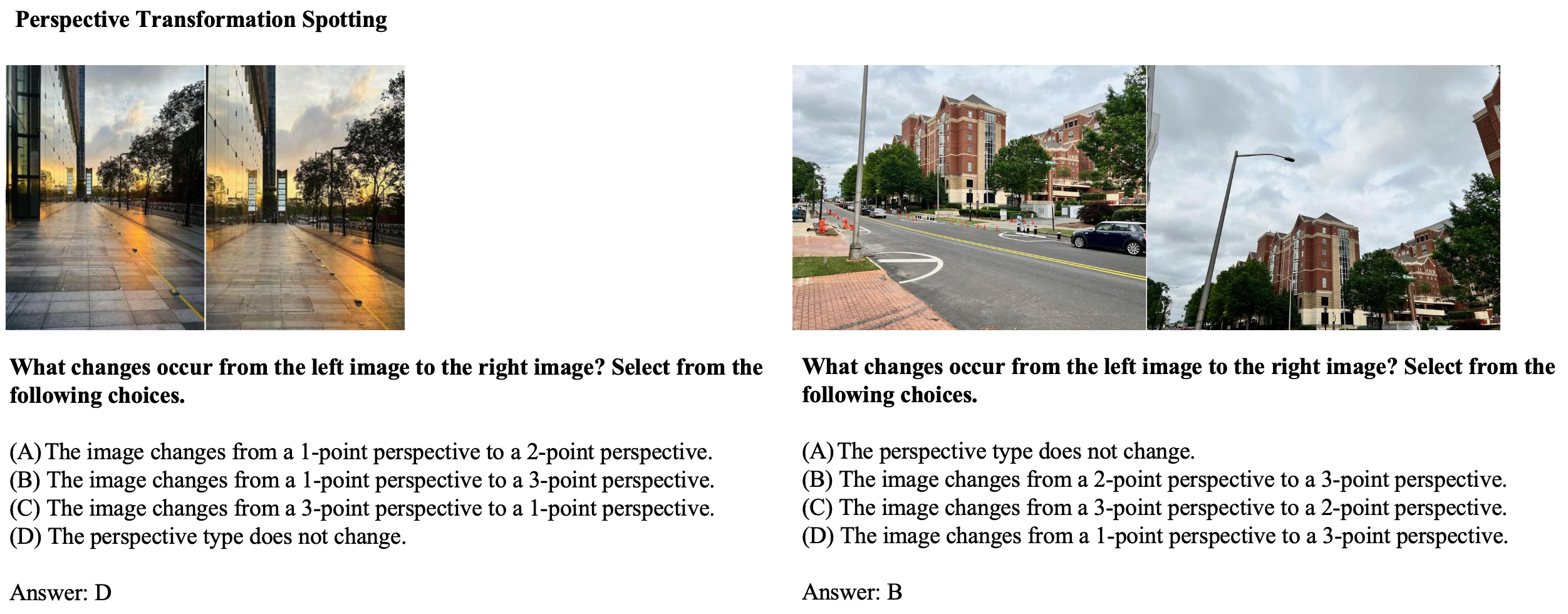}
    \caption{Examples of Perspective Transformation Spotting.}
    \label{fig:PTS}
\end{figure}
\begin{figure}
    \centering
    \includegraphics[width=1\linewidth]{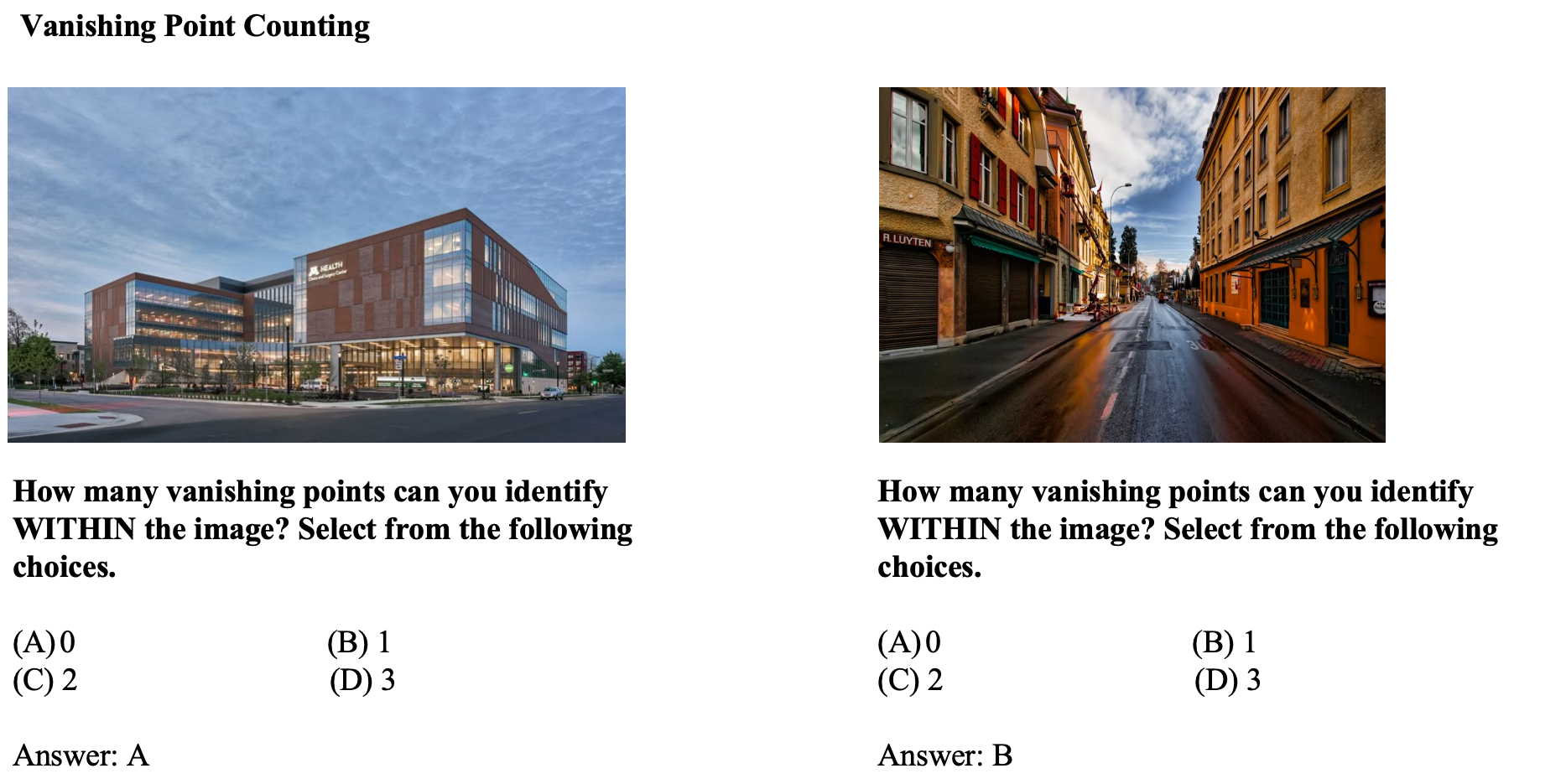}
    \caption{Examples of Vanishing Point Counting.}
    \label{fig:VPC}
\end{figure}
\begin{figure}
    \centering
    \includegraphics[width=1\linewidth]{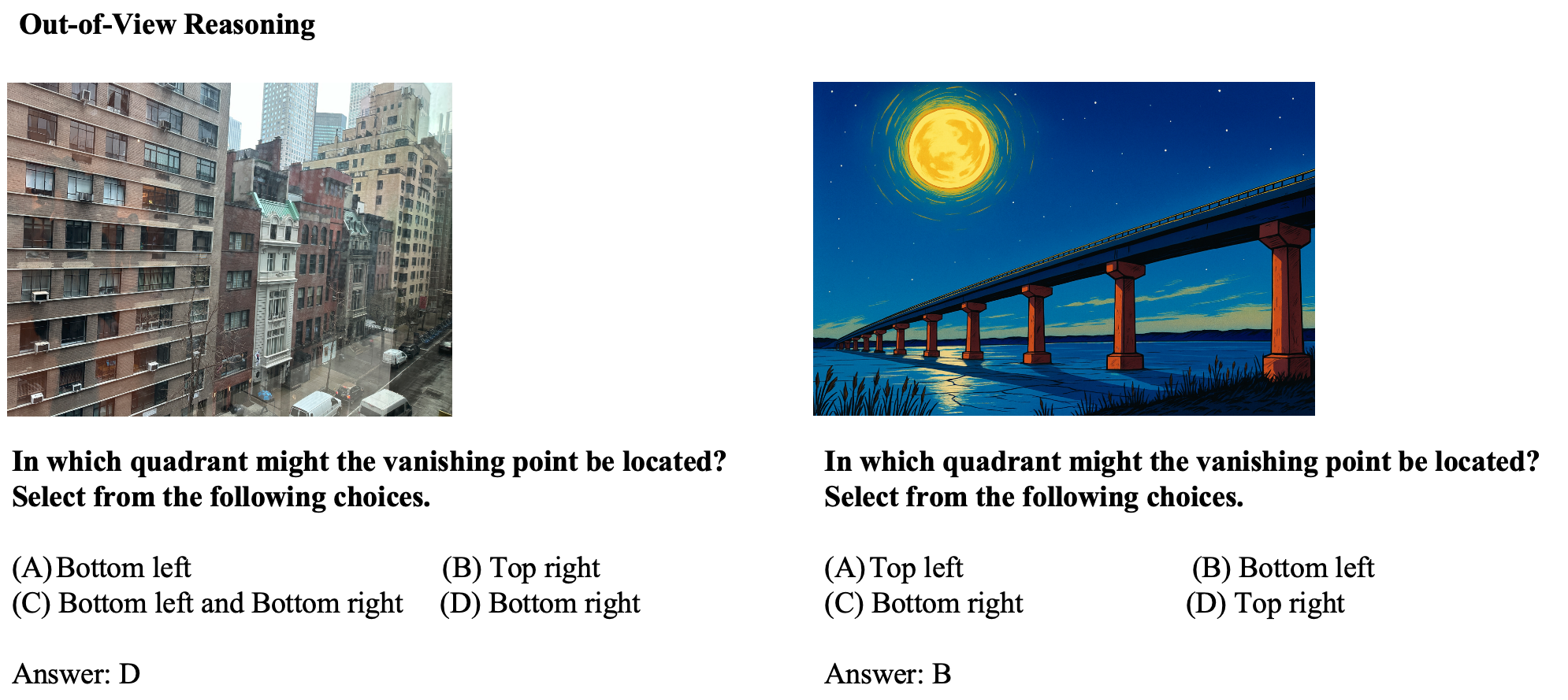}
    \caption{Examples of Out-of-View Reasoning.}
    \label{fig:OVR}
\end{figure}

\section{More Terminology of Perspective}
\begin{figure}
    \centering
    \includegraphics[width=0.87\linewidth]{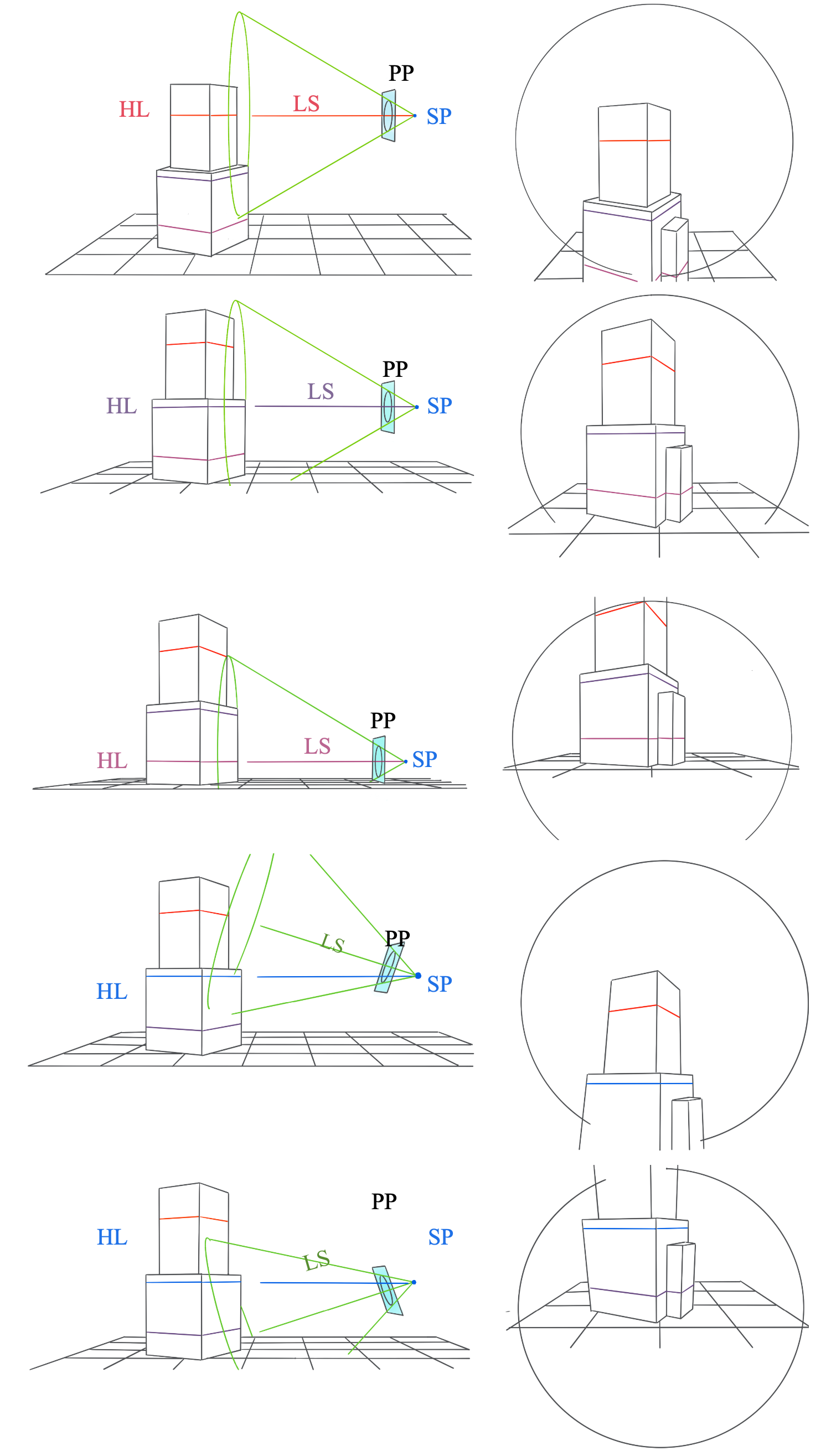}
    \caption{The relationship between Station Point (SP), Picture Plane (PP), Line of Sight (LS), and Horizon Line (HL) in perspective drawing. They demonstrate how viewing objects from different heights and angles affects spatial representation, emphasizing the critical distinction between LS and HL for accurate perspective construction. Figures are adapted from~\citep{robertson2013draw}.}
    \label{fig:LSvsHL}
\end{figure}

\Cref{fig:LSvsHL} illustrates the key distinction between the Line of Sight (LS) and Horizon Line (HL) in perspective drawing. HL represents the viewer’s eye level, while LS indicates the exact direction the viewer is looking.
When LS is parallel to the ground, it aligns with HL, resulting in a typical 2-point perspective with verticals remaining straight. But when LS tilts upward or downward, it separates from HL, introducing vertical convergence and shifting the drawing into 3-point perspective.
Importantly, the relative position of LS and HL also determines the view angle. If LS is above HL, the viewer is looking up (upward view); if it’s below, the viewer is looking down (downward view). This shift changes what parts of an object are emphasized, more base or more top, and impacts how space is perceived.

\begin{figure}
    \centering
    \includegraphics[width=1\linewidth]{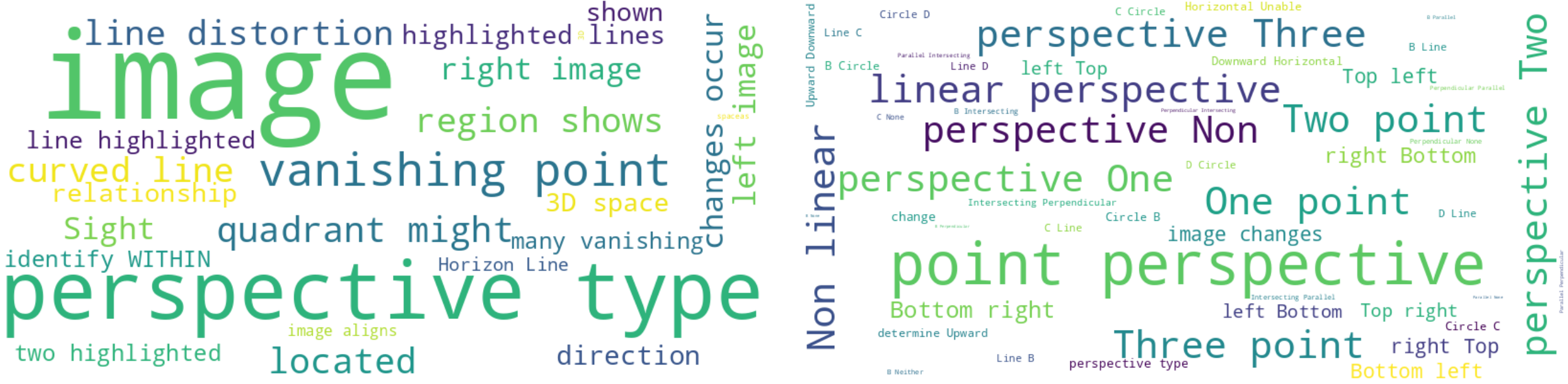}
    \caption{Word clouds of questions (left) and answer choices (right) in the MMPerspective Benchmark, illustrating the distribution of key terms related to perspective understanding.}
    \label{fig:wordcloud}
\end{figure}

\section{More Visualization}

\subsection{Model Size \& Performance for Each Task}
\label{appendix:vlm_perf_tasks}
\begin{figure}[!ht]
\centering

    \begin{minipage}[b]{0.32\linewidth}
        \centering
        \includegraphics[width=\linewidth]{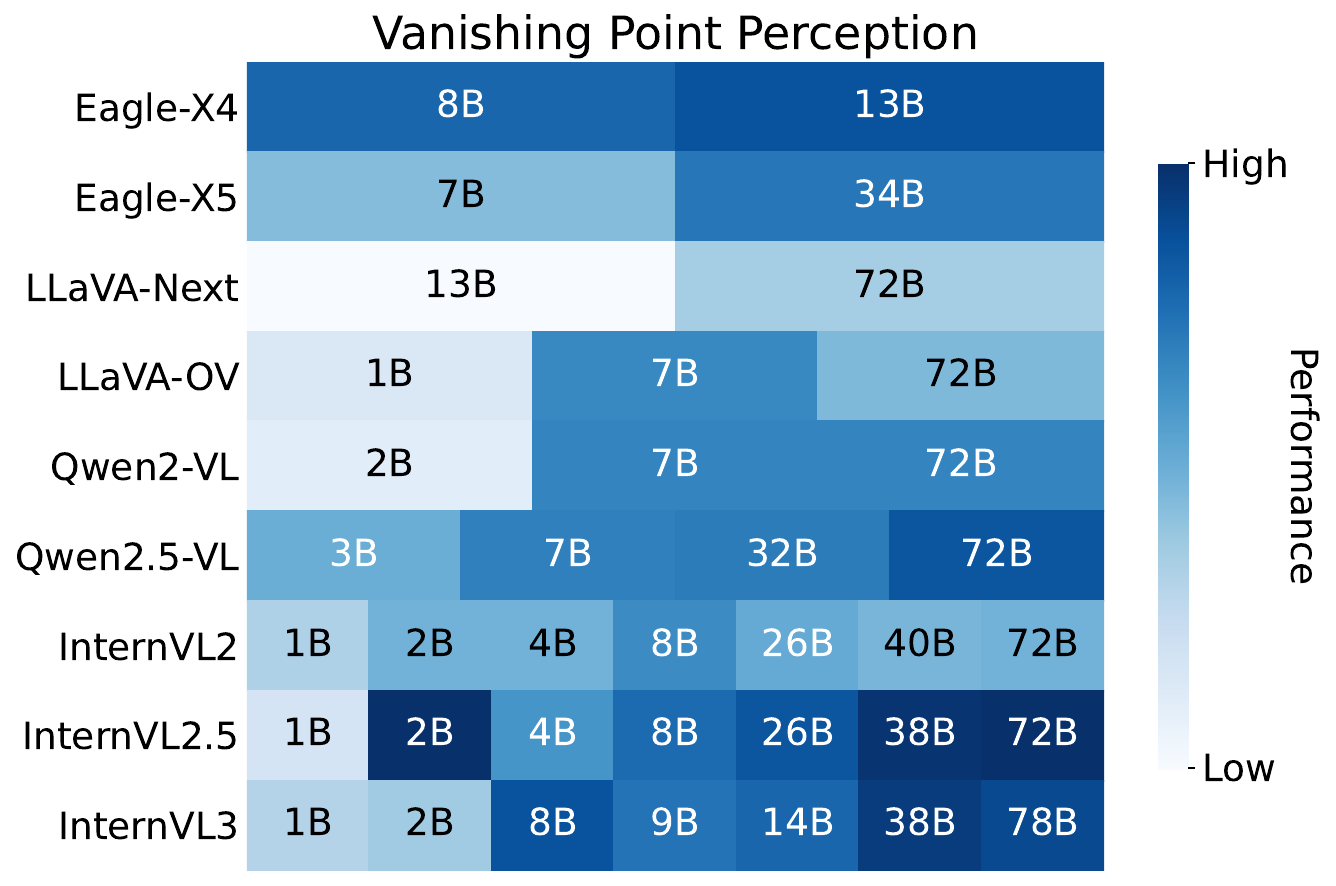}
        \caption{The heatmap for Vanishing Point Perception.}
        \label{fig:heatmap_vanishing_point}
    \end{minipage}
    \hfill
    \begin{minipage}[b]{0.32\linewidth}
        \centering
        \includegraphics[width=\linewidth]{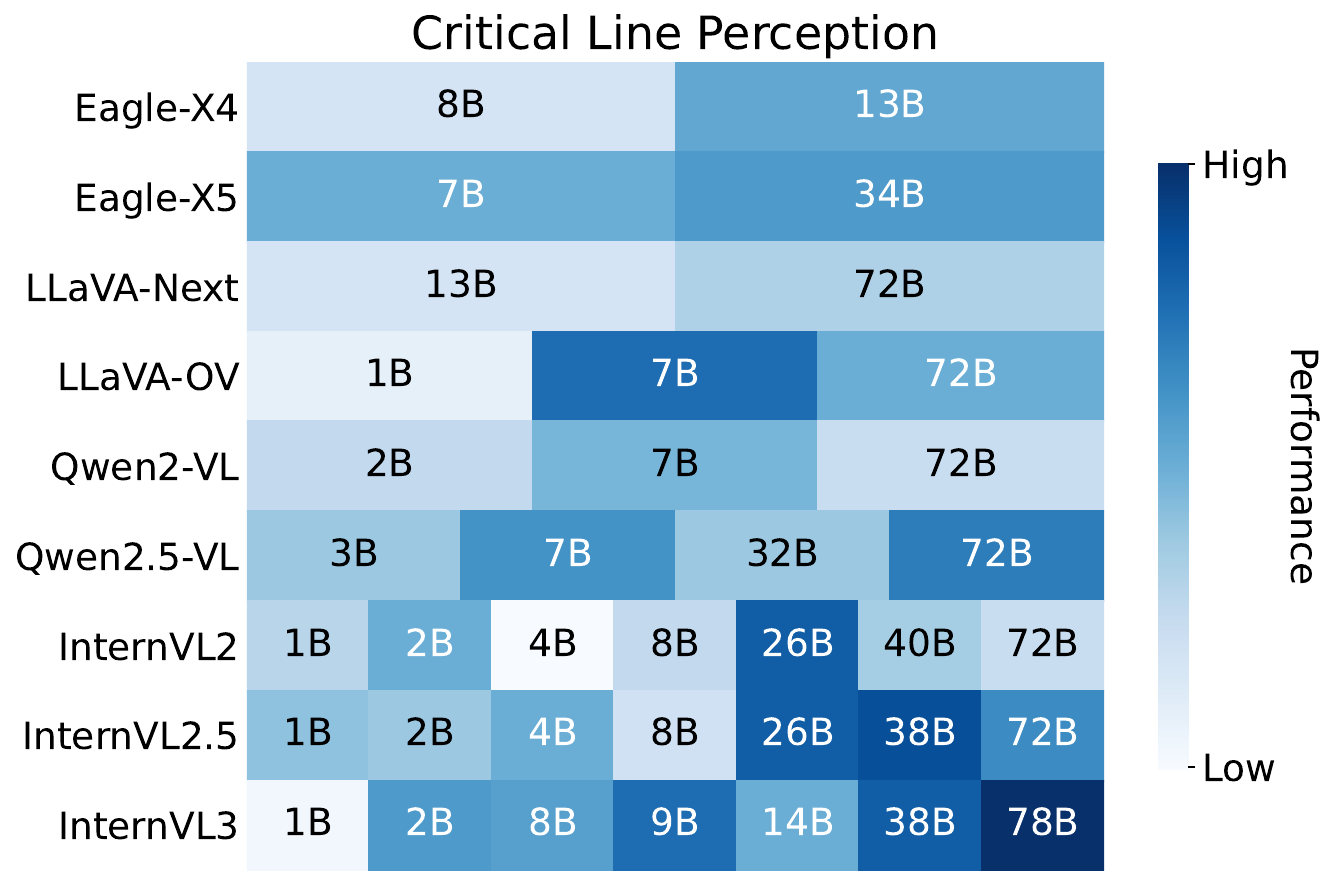}
        \caption{The heatmap for Critical Line Perception.}
        \label{fig:heatmap_critical_line}
    \end{minipage}
    \hfill
    \begin{minipage}[b]{0.32\linewidth}
        \centering
        \includegraphics[width=\linewidth]{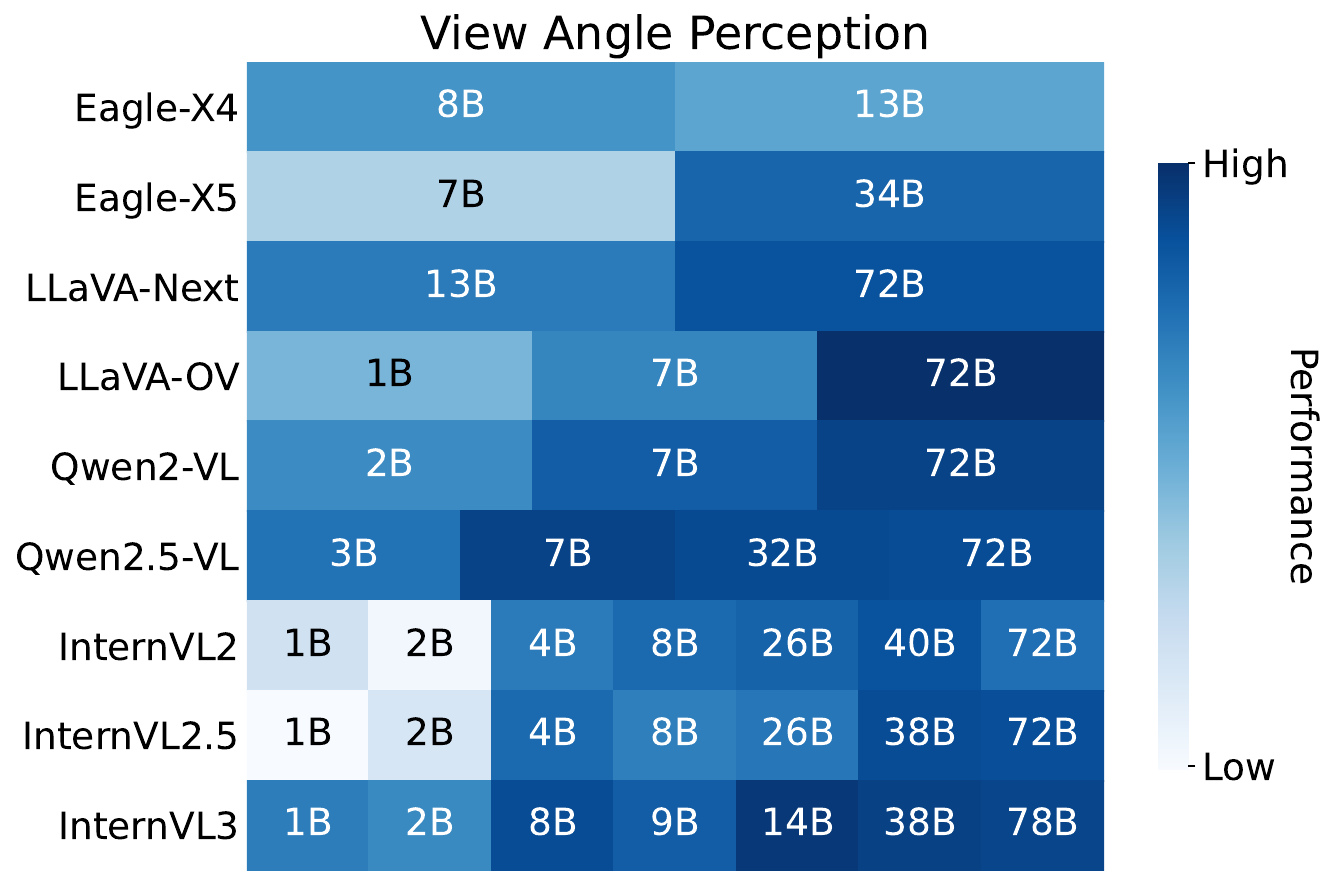}
        \caption{The heatmap for View Angle Perception.}
        \label{fig:heatmap_view_angle}
    \end{minipage}

\end{figure}

\begin{figure}[!ht]
\vspace{-4mm}
\centering

    \begin{minipage}[b]{0.32\linewidth}
        \centering
        \includegraphics[width=\linewidth]{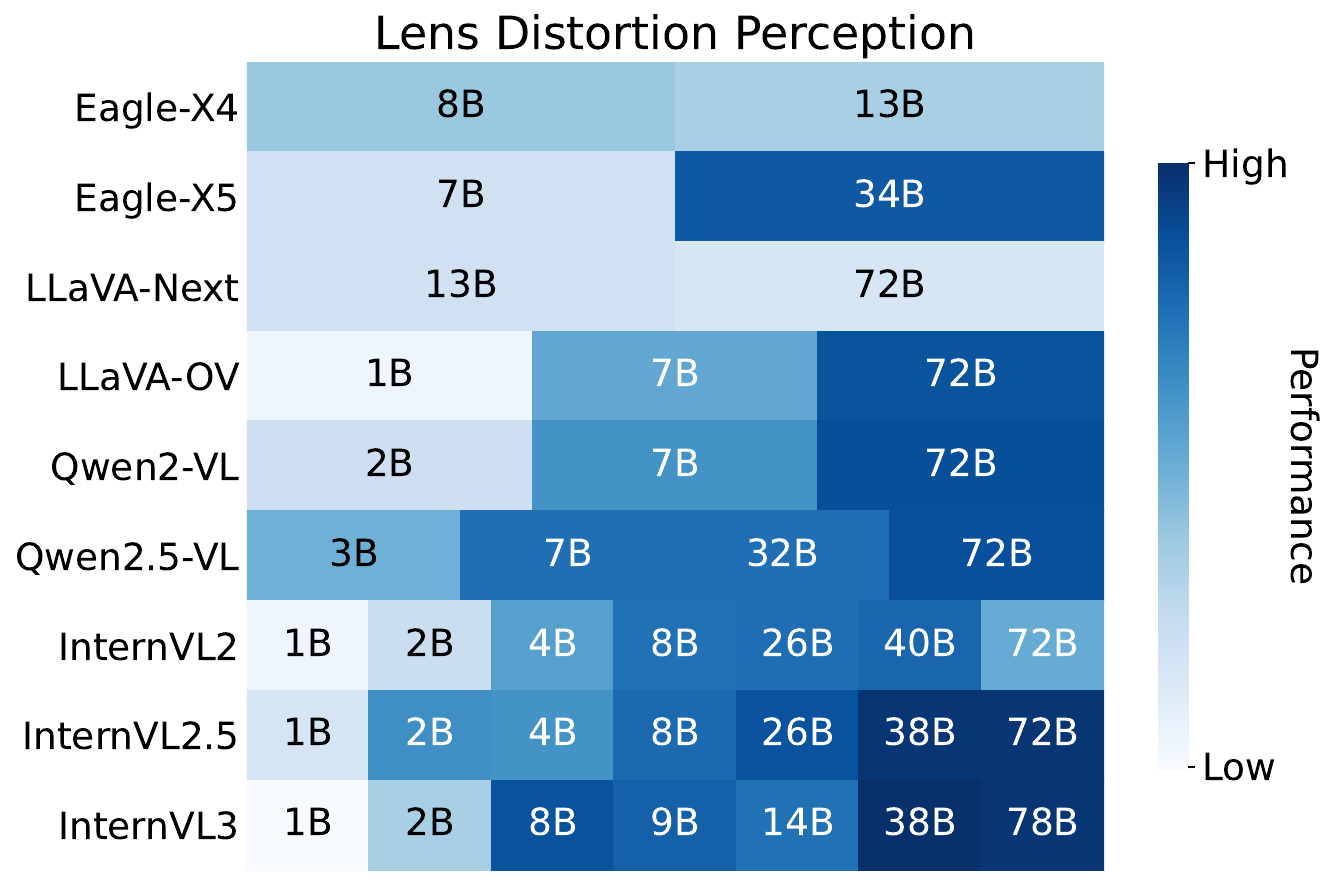}
        \caption{The heatmap for Lens Distortion Perception.}
        \label{fig:heatmap_lens_distortion}
    \end{minipage}
    \hfill
    \begin{minipage}[b]{0.32\linewidth}
        \centering
        \includegraphics[width=\linewidth]{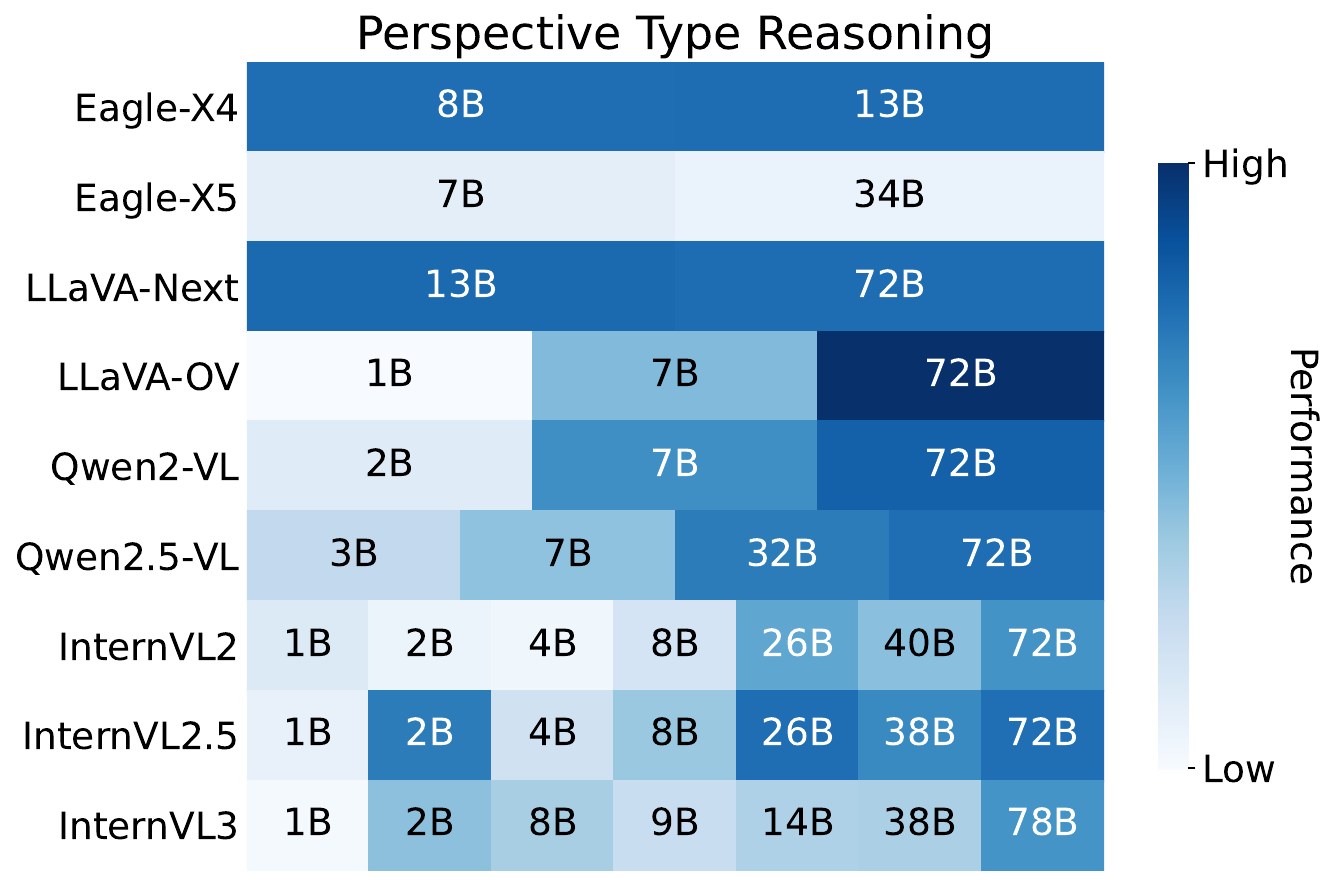}
        \caption{The heatmap for Perspective Type Reasoning.}
        \label{fig:heatmap_perspective_type}
    \end{minipage}
    \hfill
    \begin{minipage}[b]{0.32\linewidth}
        \centering
        \includegraphics[width=\linewidth]{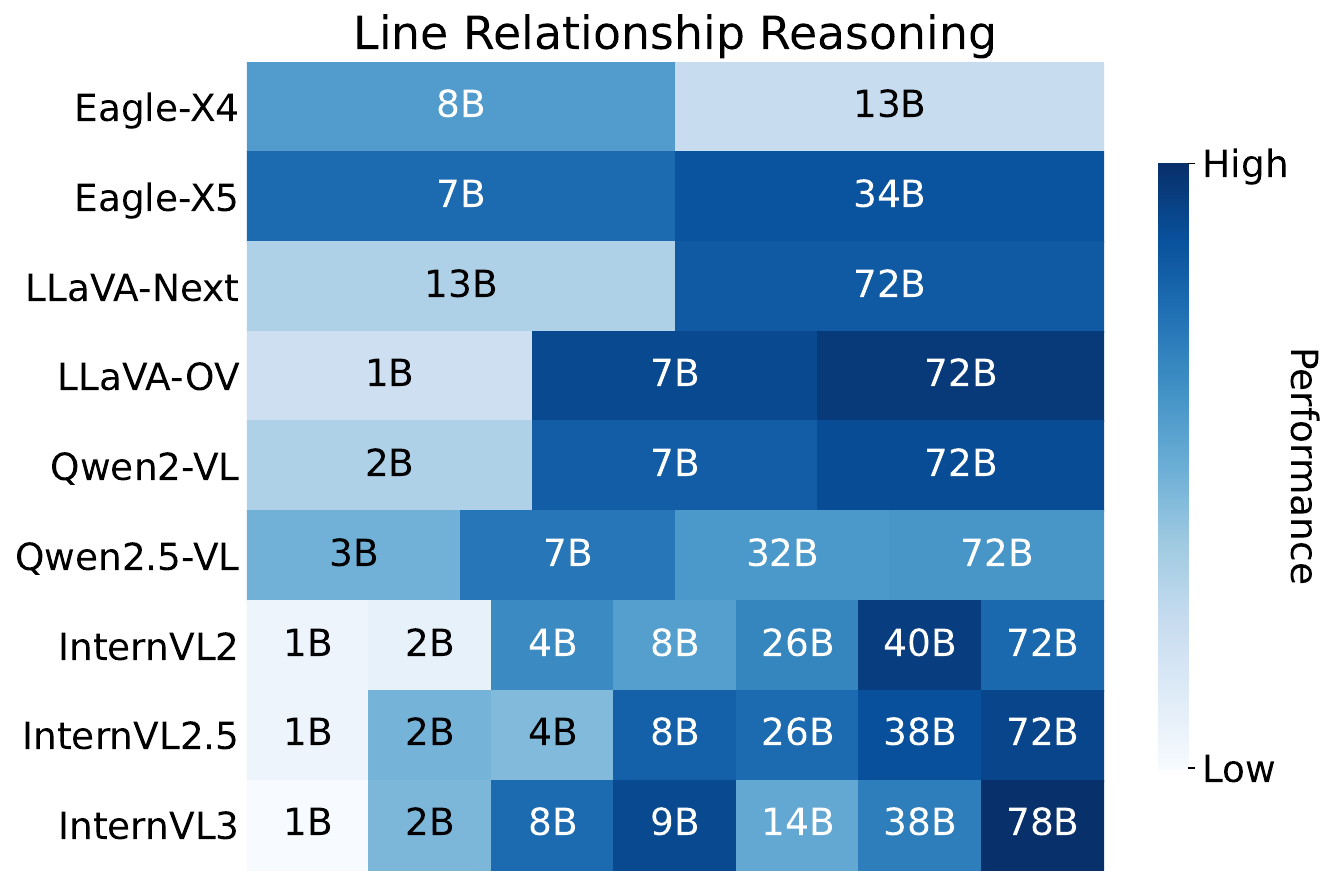}
        \caption{The heatmap for Line Relationship Reasoning.}
        \label{fig:heatmap_line_relationship}
    \end{minipage}

\end{figure}

\begin{figure}[!ht]
\vspace{-4mm}
\centering

    \begin{minipage}[b]{0.32\linewidth}
        \centering
        \includegraphics[width=\linewidth]{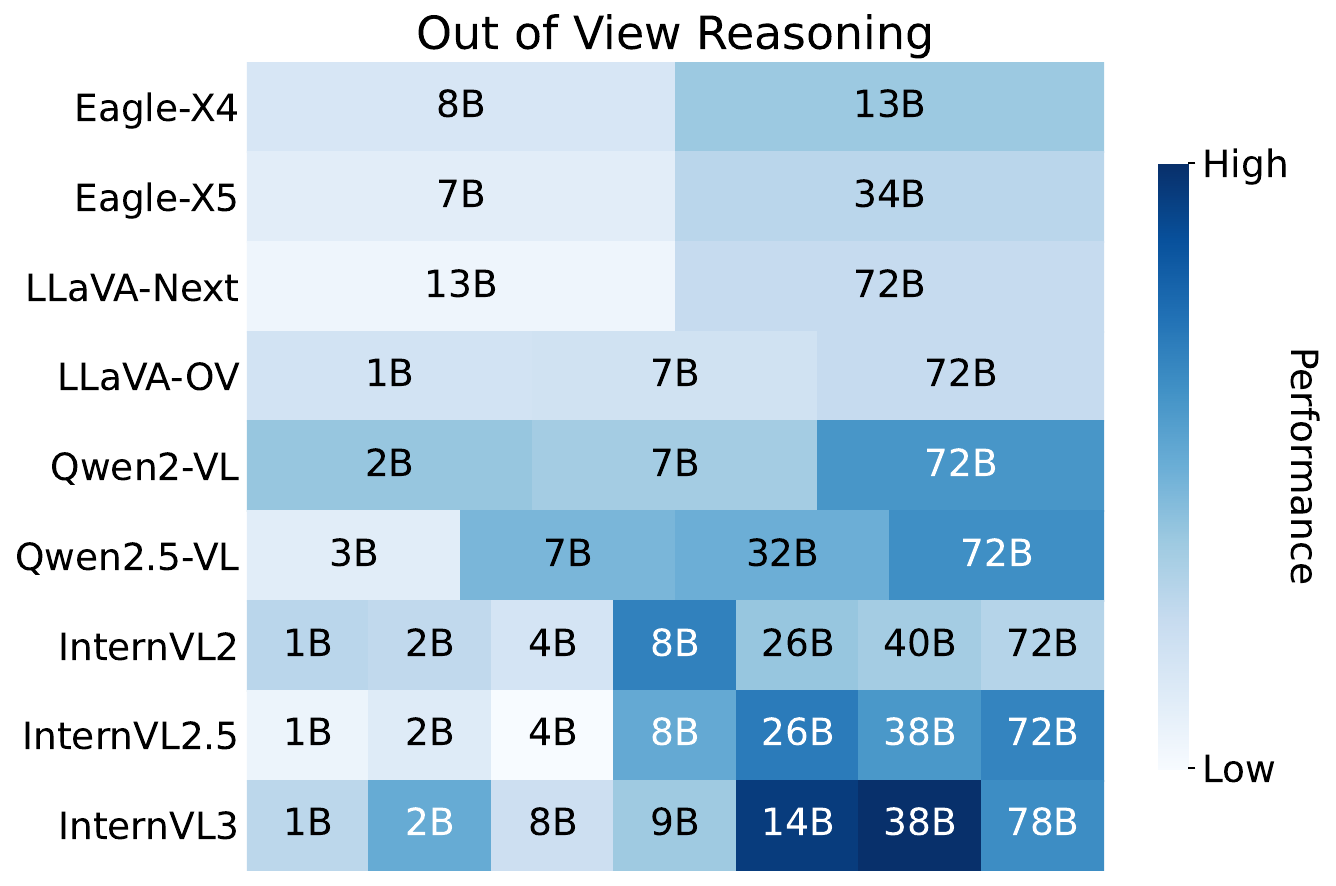}
        \caption{The heatmap for Out of View Reasoning.}
        \label{fig:heatmap_out_of_view}
    \end{minipage}
    \hfill
    \begin{minipage}[b]{0.32\linewidth}
        \centering
        \includegraphics[width=\linewidth]{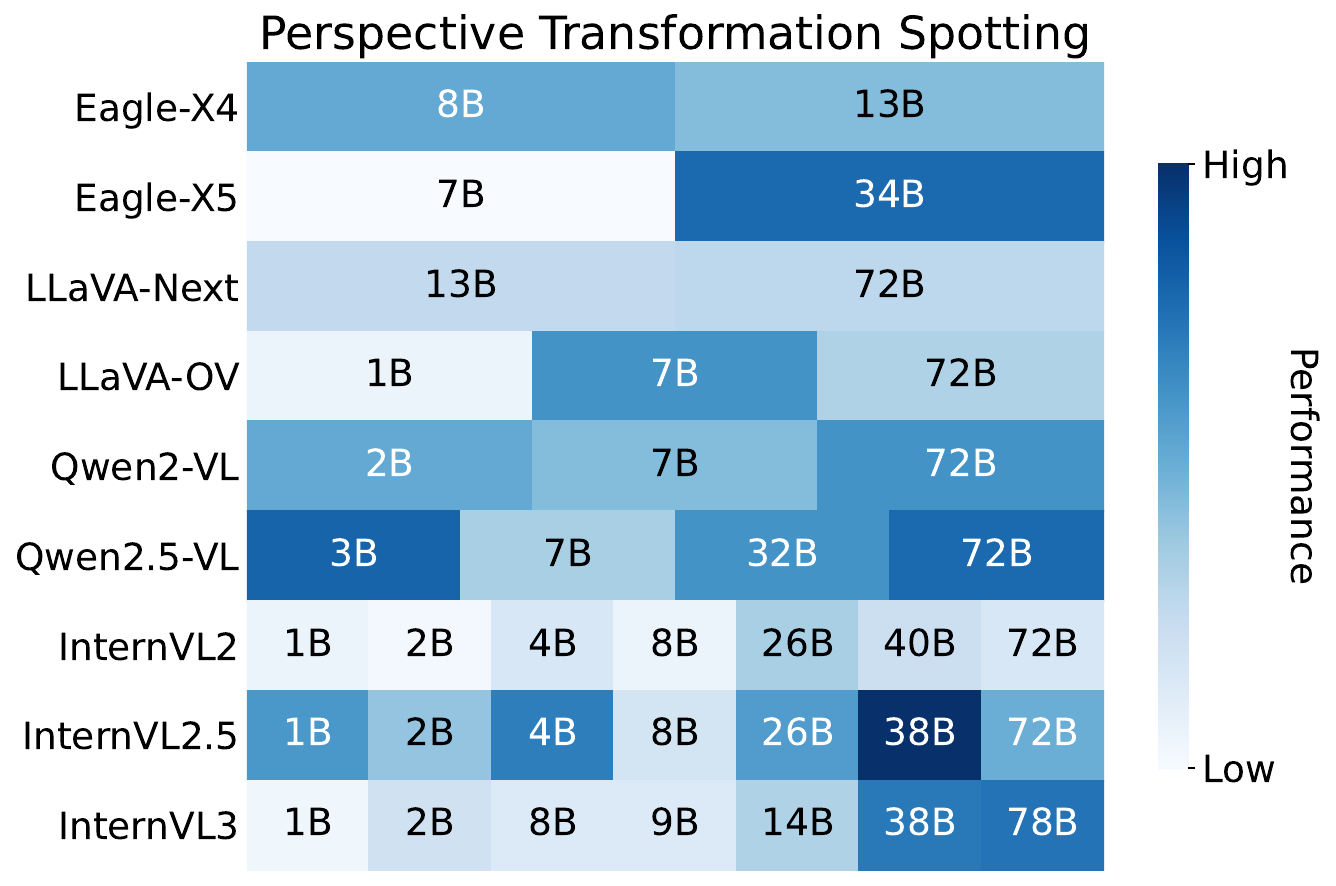}
        \caption{The heatmap for Perspective Transformation Spotting.}
        \label{fig:heatmap_perspective_transformation}
    \end{minipage}
    \hfill
    \begin{minipage}[b]{0.32\linewidth}
        \centering
        \includegraphics[width=\linewidth]{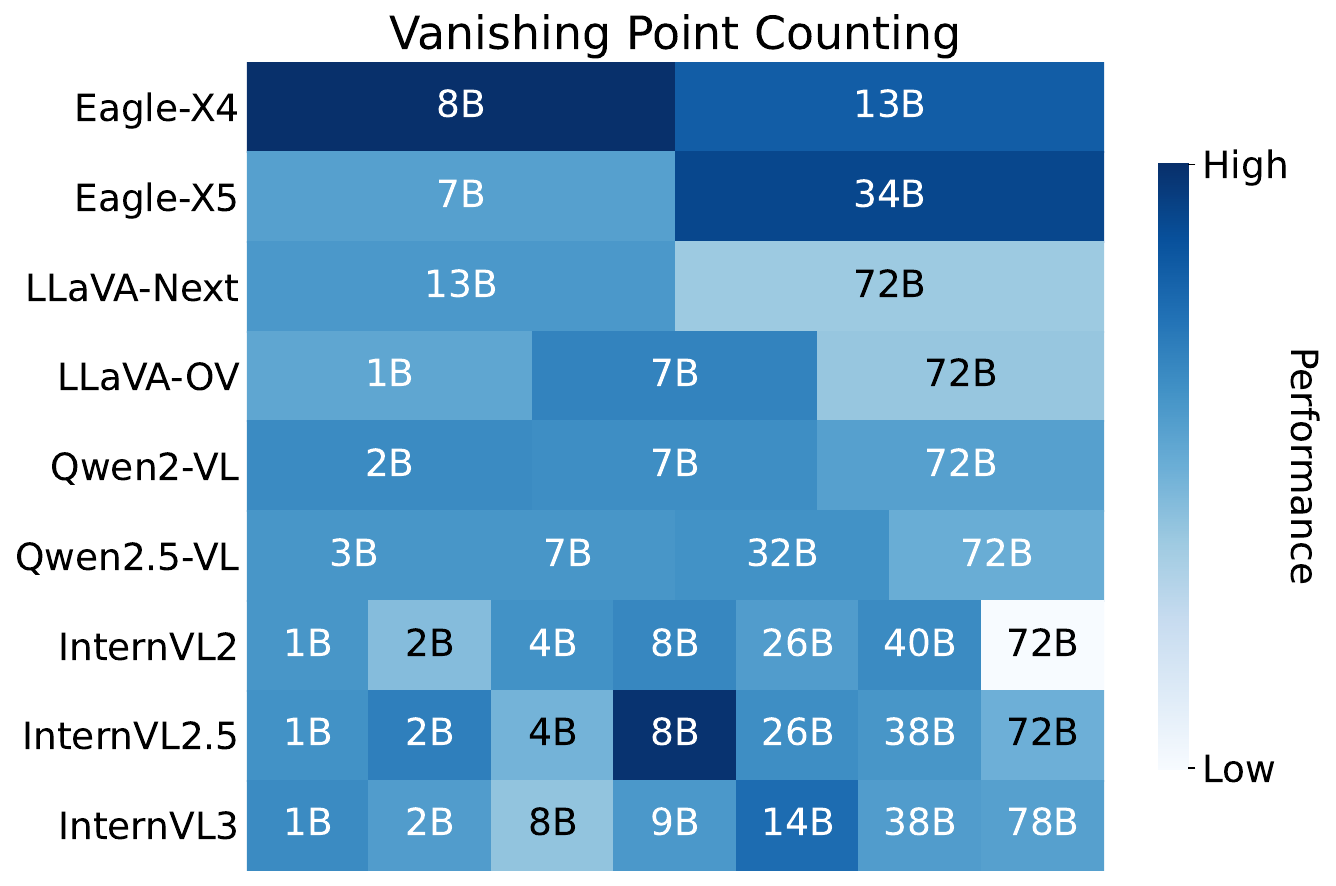}
        \caption{The heatmap for Vanishing Point Counting.}
        \label{fig:heatmap_vanishing_point_counting}
    \end{minipage}

\end{figure}

In Figure~\ref{fig:heatmap_vanishing_point} to \ref{fig:heatmap_vanishing_point_counting}, we present heatmaps for the $10$ tasks in our \textbf{\textcolor{my_red}{Perspective Perception}}, \textbf{\textcolor{my_blue}{Perspective Reasoning}}. The figures show the correlations between the sizes of \textbf{model parameters} and the metrics. Deeper color represents better performance. Each row represents a model family with the sizes growing from small to large. 
Most tasks clearly exhibit the correlation between model sizes and performance, i.e., larger model leads to higher metrics. However, Figure~\ref{fig:heatmap_perspective_transformation} shows that models with median size have better performance than smaller and larger models in \textbf{Perspective Transformation Spotting (\textcolor{my_blue}{PTS})}. Moreover, in \textbf{Vanishing Point Counting (\textcolor{my_blue}{VPC})}, we observe a reversed correlation where larger models lead to worse performance.




\subsection{Effect of Chain-of-Thought}
\begin{figure}
    \centering
    \includegraphics[width=1\linewidth]{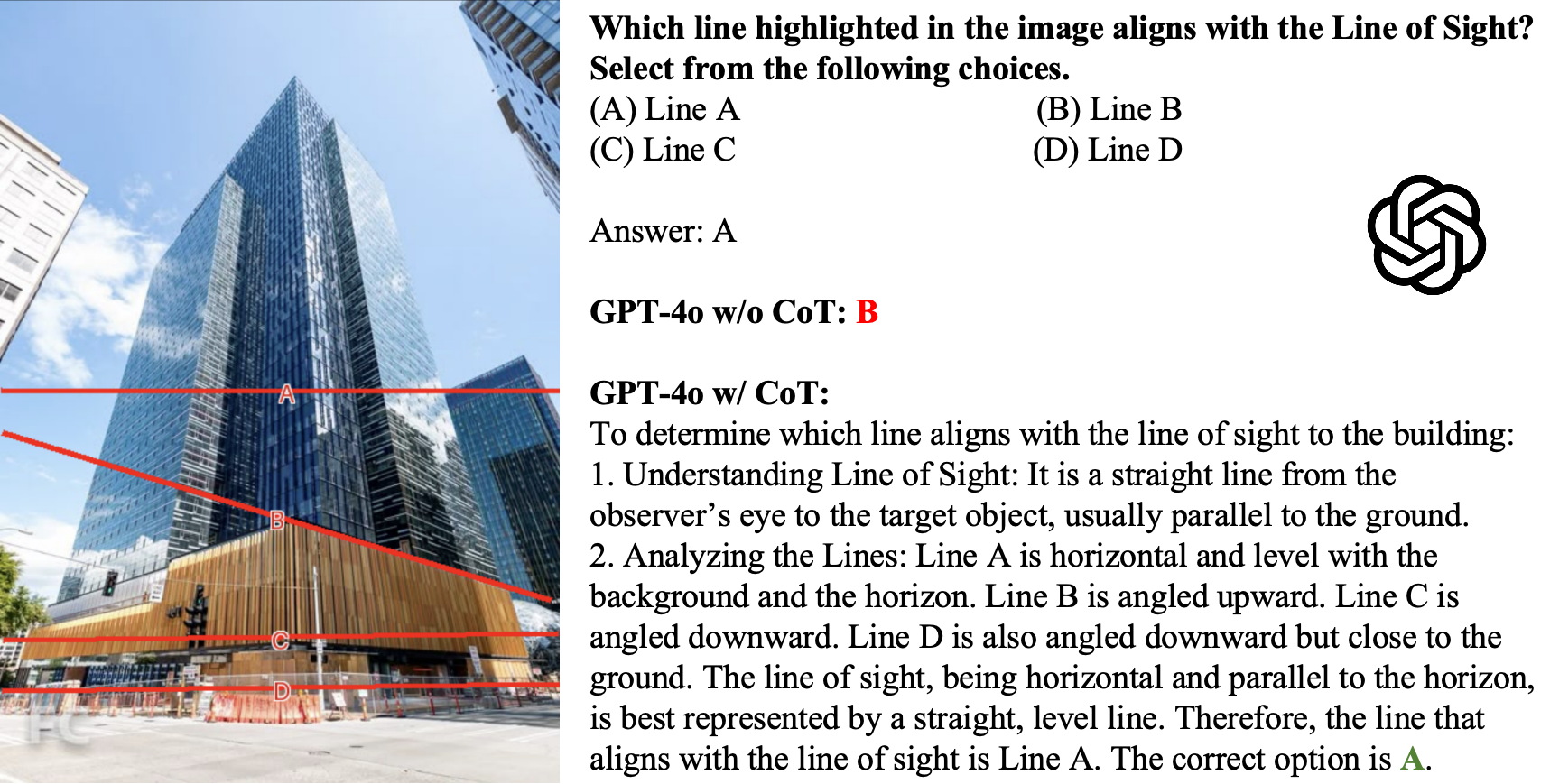}
    \caption{Examples of Chain-of-Thought Reasoning.}
    \label{fig:cot_gpt_1}
\end{figure}

\begin{figure}
    \centering
    \includegraphics[width=1\linewidth]{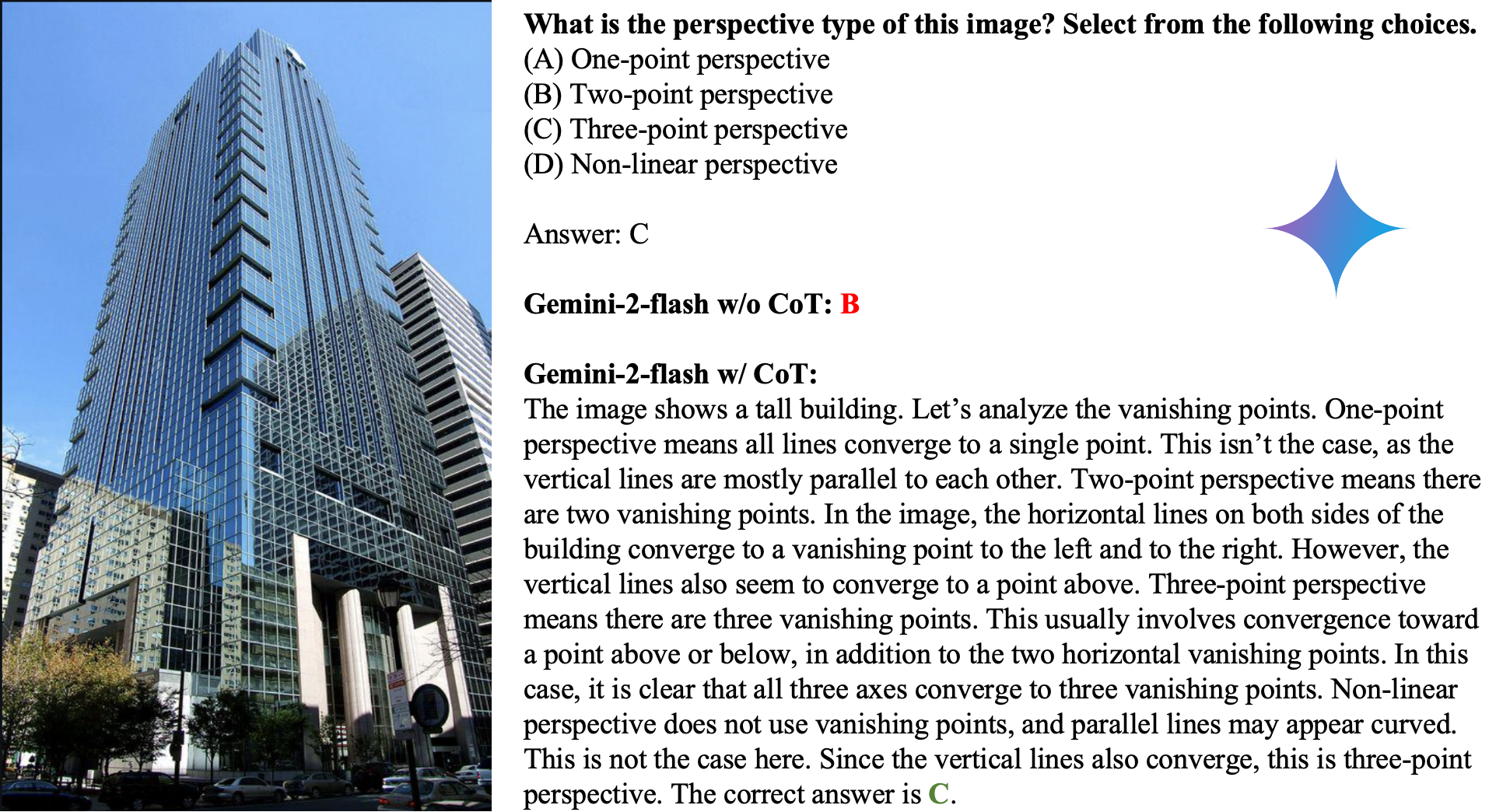}
    \caption{Examples of Chain-of-Thought Reasoning.}
    \label{fig:cot_gemini_1}
\end{figure}

\begin{figure}
    \centering
    \includegraphics[width=1\linewidth]{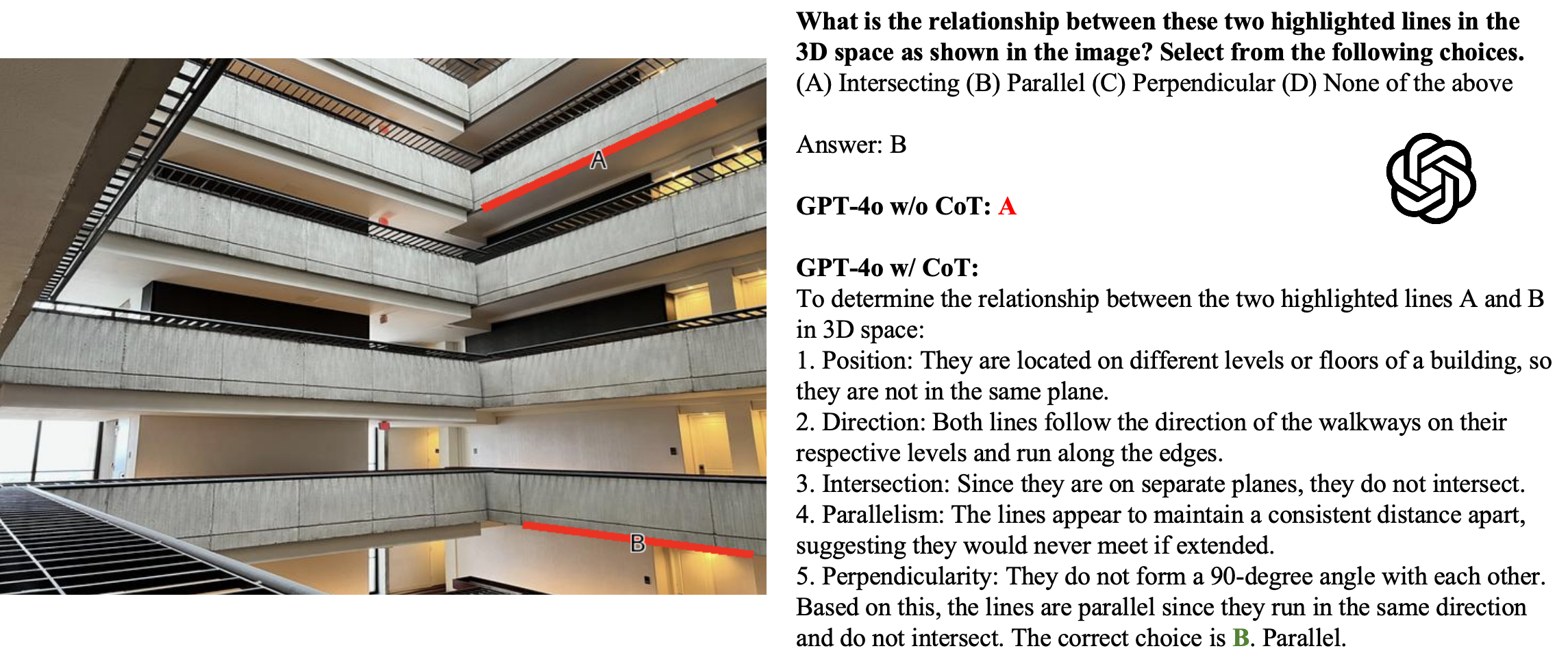}
    \caption{Examples of Chain-of-Thought Reasoning.}
    \label{fig:cot_gpt_2}
\end{figure}

\begin{figure}
    \centering
    \includegraphics[width=1\linewidth]{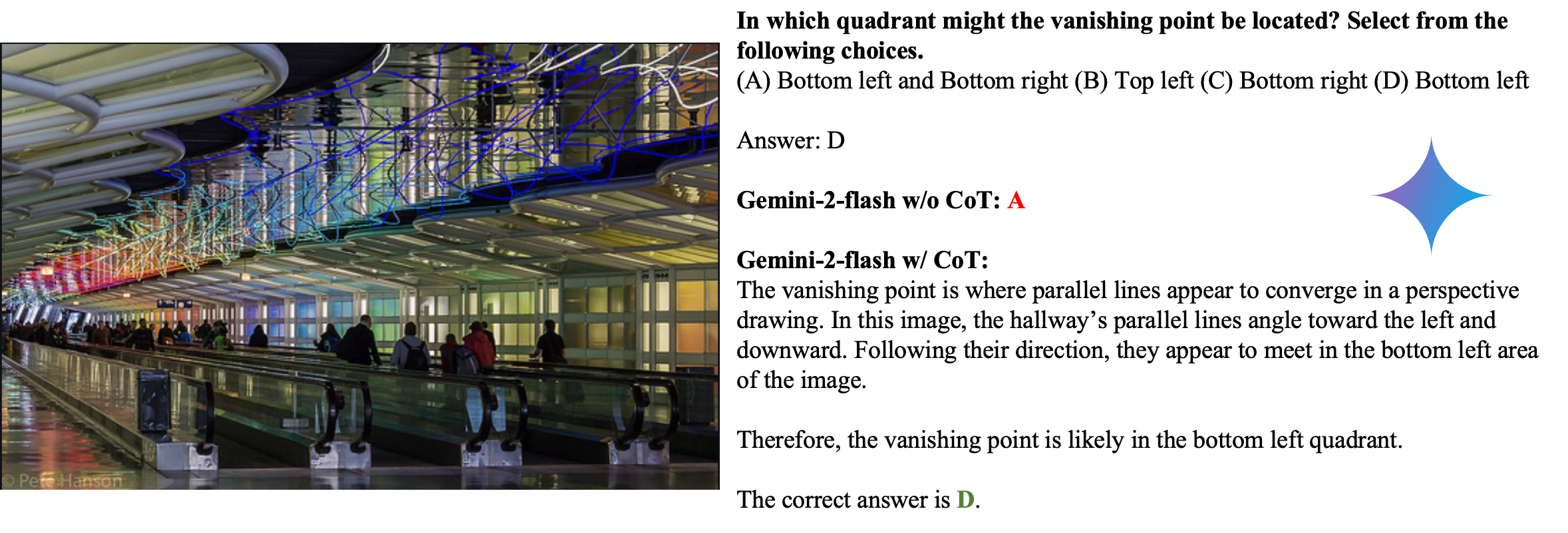}
    \caption{Examples of Chain-of-Thought Reasoning.}
    \label{fig:cot_gemini_2}
\end{figure}

Figures~\ref{fig:cot_gpt_1} to \ref{fig:cot_gemini_2} are examples that demonstrate how Chain-of-Thought (CoT) can generally enhance the model's performance.

\begin{figure}
    \centering
    \includegraphics[width=1\linewidth]{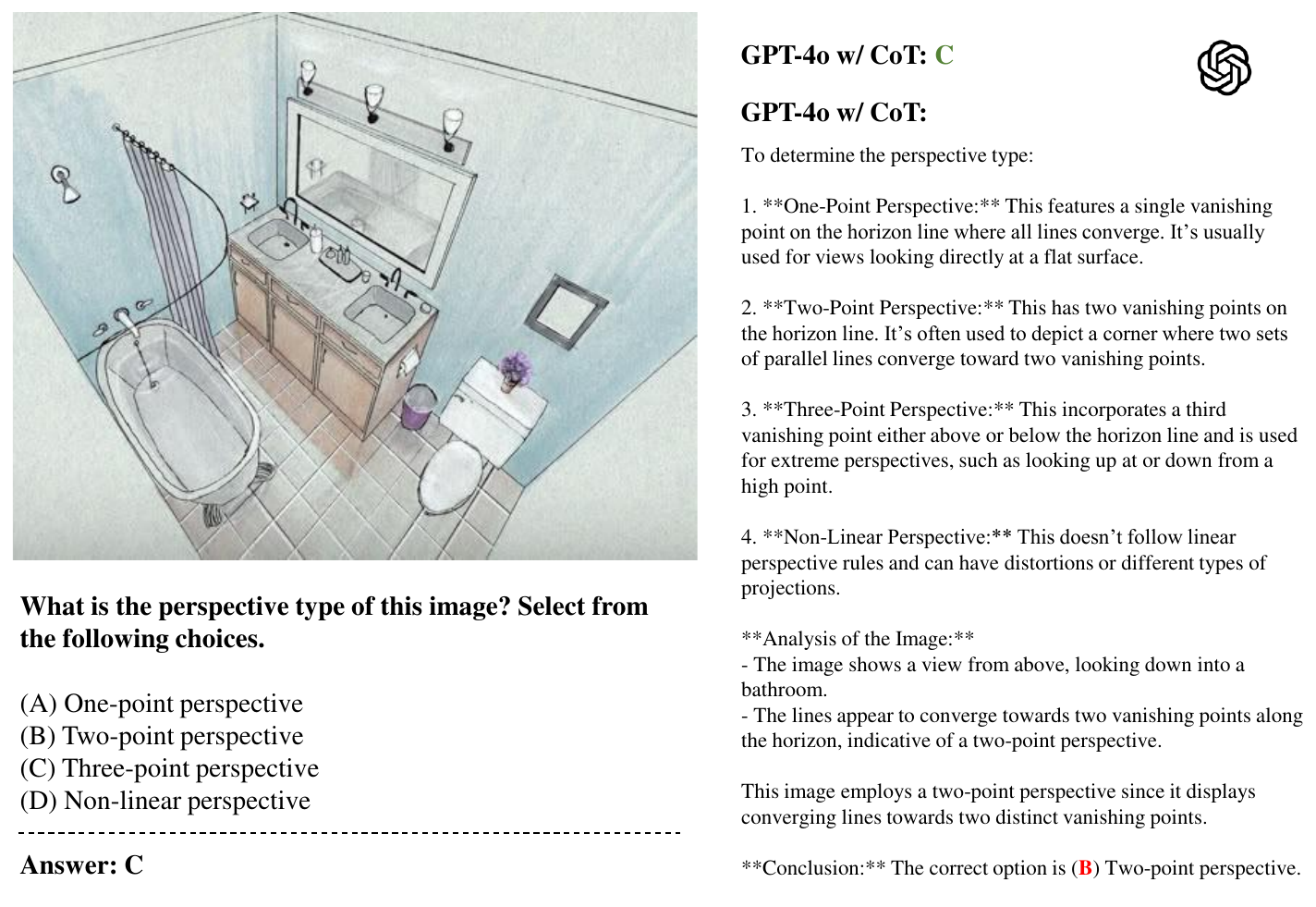}
    \caption{Examples of GPT-4o with Chain-of-Thought Reasoning in Perspective Type Reasoning.}
    \label{fig:cot_failure_gpt4o_ptr}
\end{figure}

\begin{figure}
    \centering
    \includegraphics[width=1\linewidth]{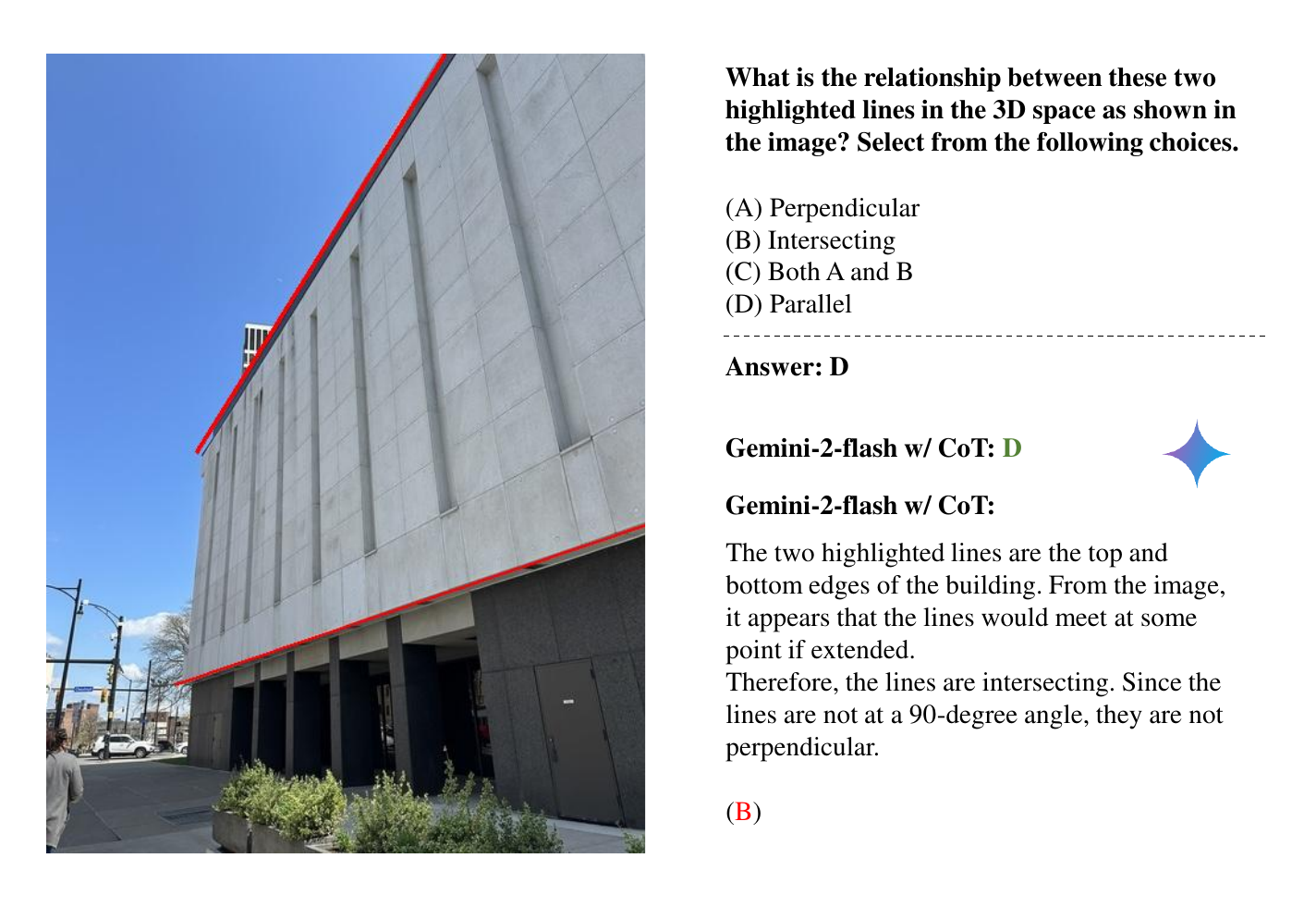}
    \caption{Examples of Gemini-2-flash with Chain-of-Thought Reasoning in Perspective Transformation Spotting.}
    \label{fig:cot_failure_gemini2_pts}
\end{figure}

\begin{figure}
    \centering
    \includegraphics[width=1\linewidth]{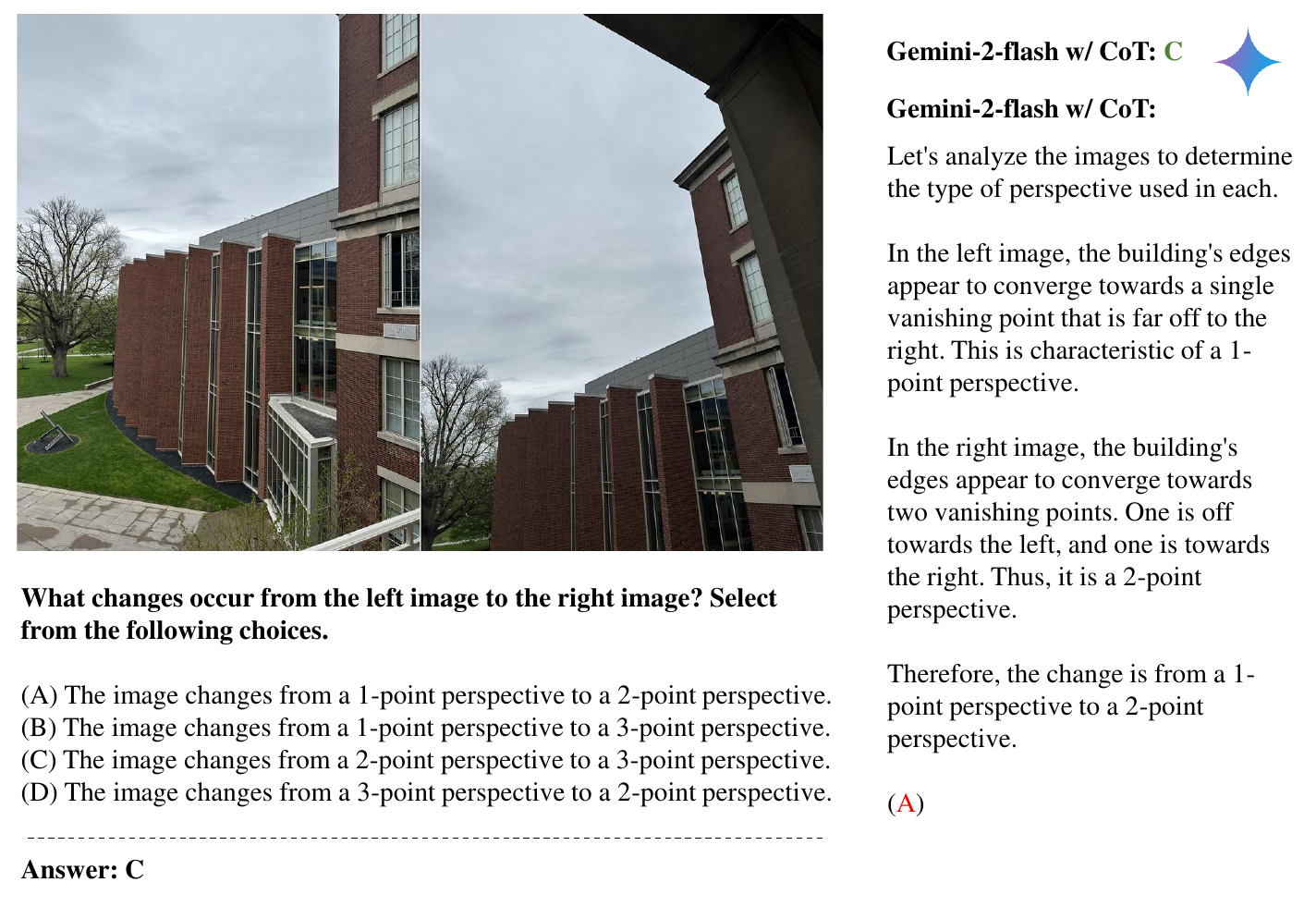}
    \caption{Examples of Gemini-2-flash with Chain-of-Thought Reasoning in Line Relationship Reasoning.}
    \label{fig:cot_failure_gemini2_lrr}
\end{figure}

Despite the general enhancement, a few failures still emerge. Figures~\ref{fig:cot_failure_gpt4o_ptr}, \ref{fig:cot_failure_gemini2_pts}, and \ref{fig:cot_failure_gemini2_lrr} show three representative failure cases, including GPT-4o on Perspective Type Reasoning, and Gemini-2-flash on Line Relationship and Perspective Transformation Spotting. In these three cases, we all observed that the models made direct factual errors when analyzing the information in the images, rather than logical errors during the CoT process. This indicates that what limits the performance of the model is the ability to understand images.


\subsection{Performance for Each Model Family}
\begin{figure}
    \centering
    \includegraphics[width=0.9\linewidth]{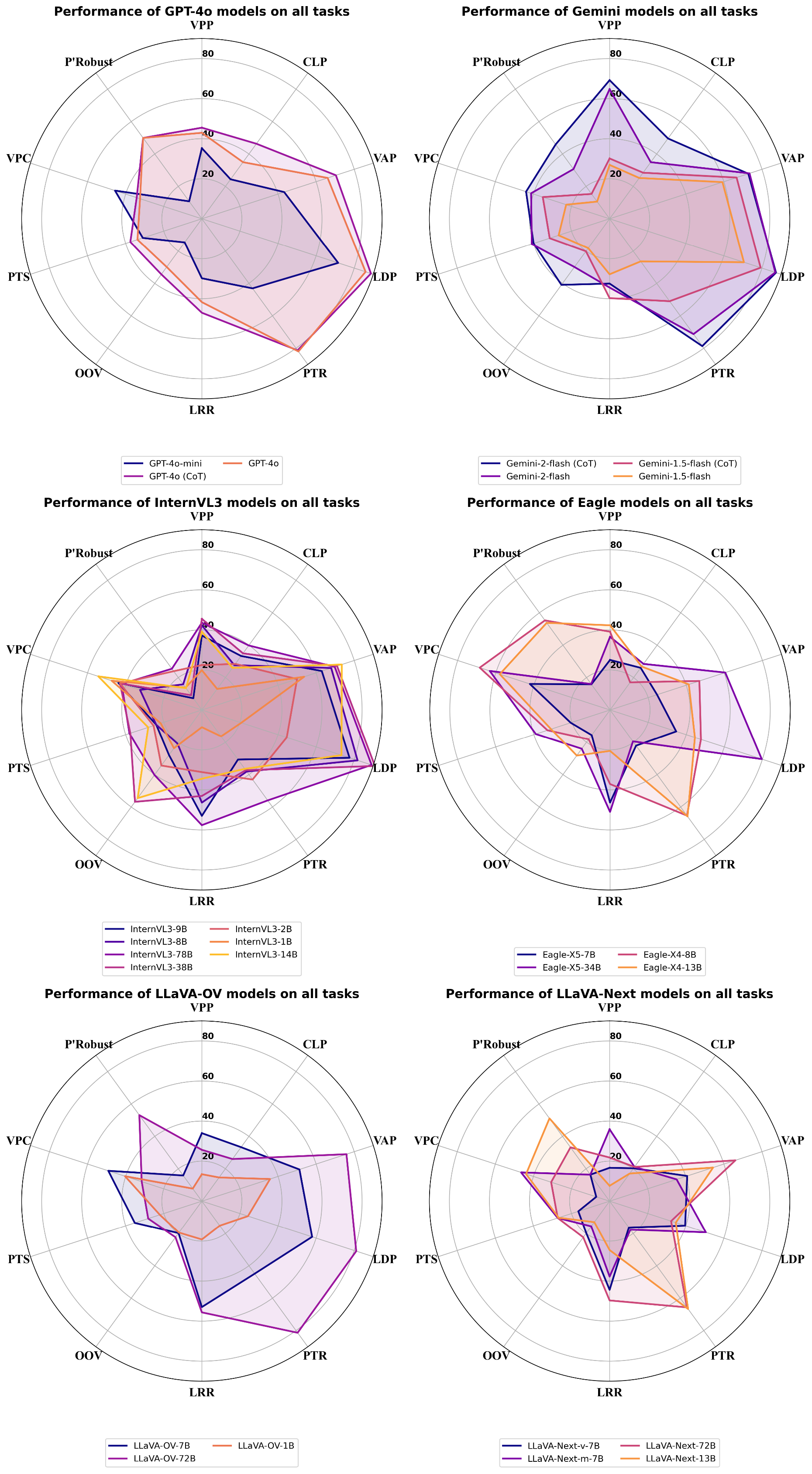}
\caption{Task performance of models within each family (part 1).}
    \label{fig:radar_1}
\end{figure}

\begin{figure}
    \centering
\includegraphics[width=0.9\linewidth]{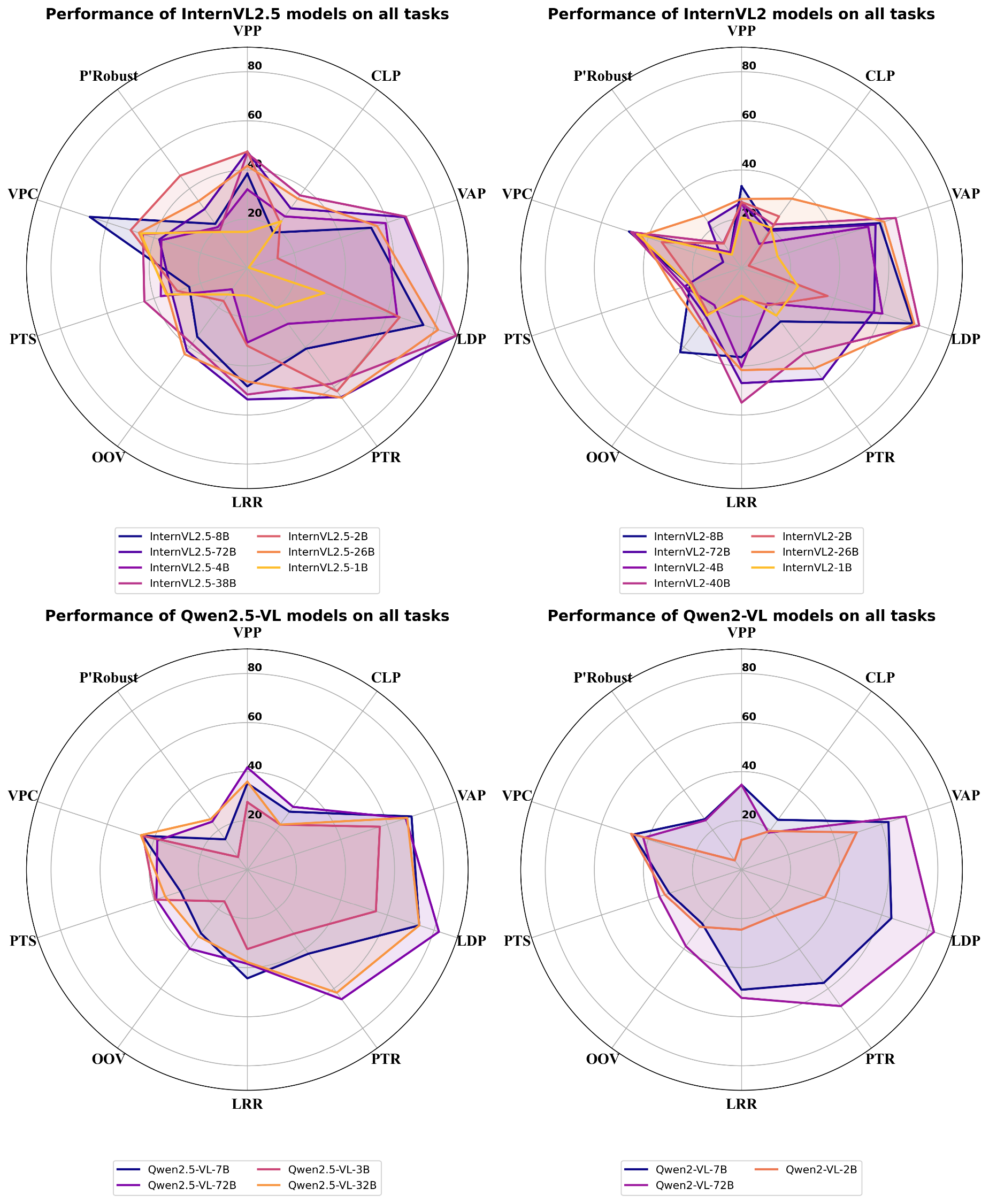}
\caption{Task performance of models within each family (part 2).}
    \label{fig:radar_2}
\end{figure}

\Cref{fig:radar_1} and \Cref{fig:radar_2} show task performance across various models within the same model families. Generally, models that are larger usually excel in most tasks.

\subsection{Question Difficulty Distribution}
\begin{figure}
    \centering
    \includegraphics[width=0.95\linewidth]{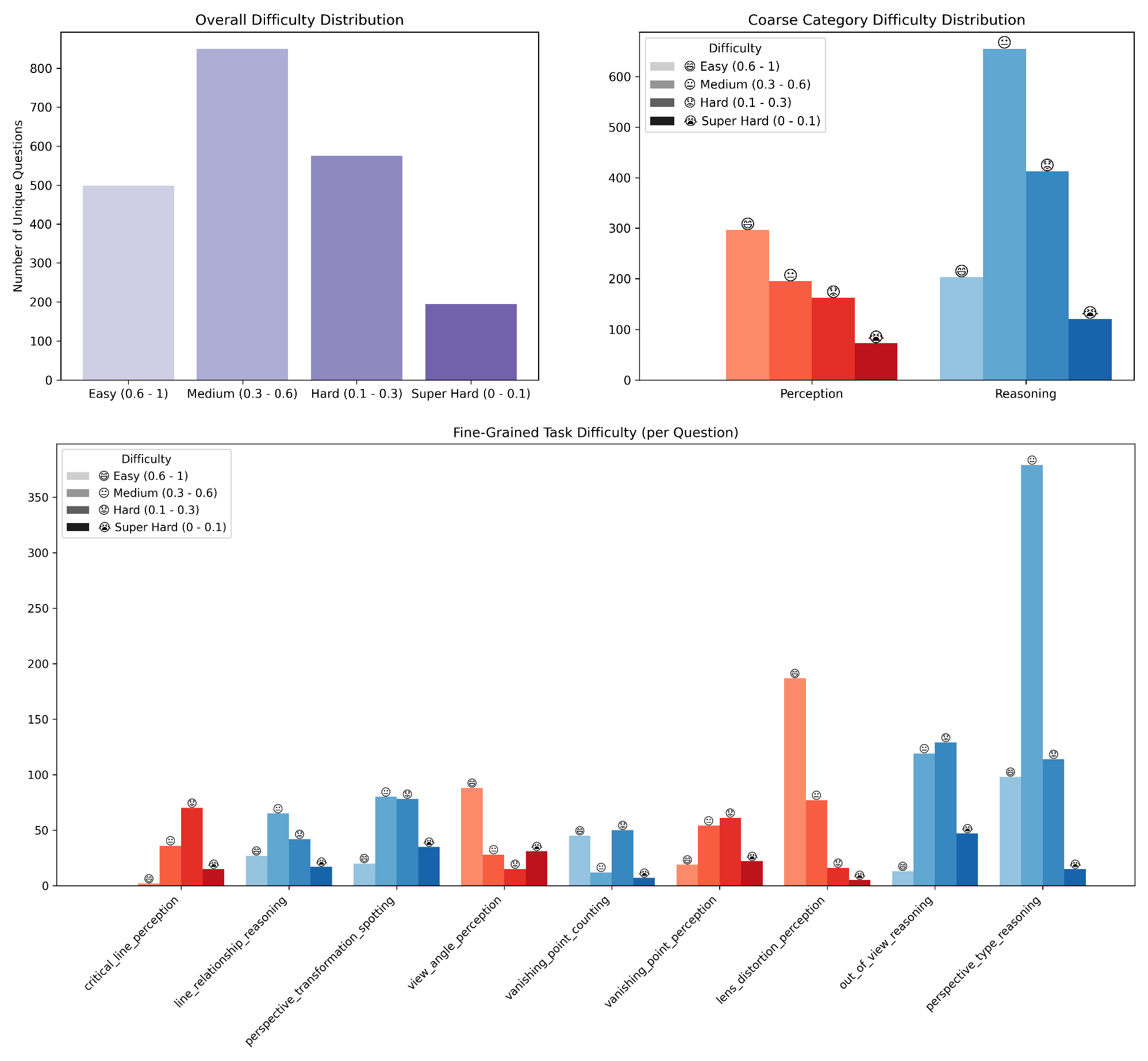}
\caption{Distribution of question difficulty across task types.}
    \label{fig:diff}
\end{figure}

\Cref{fig:diff} presents the question difficulty distribution based on average model accuracy. Each question is categorized into four difficulty levels, \textit{Easy}, \textit{Medium}, \textit{Hard}, and \textit{Super Hard}, based on the proportion of models that answered it correctly.
The top two charts show the overall and type-level distributions, while the bottom figure provides a fine-grained view across tasks.

\section{Annotation}
\subsection{Annotation Tool}
\begin{figure}
    \centering
    \includegraphics[width=0.9\linewidth]{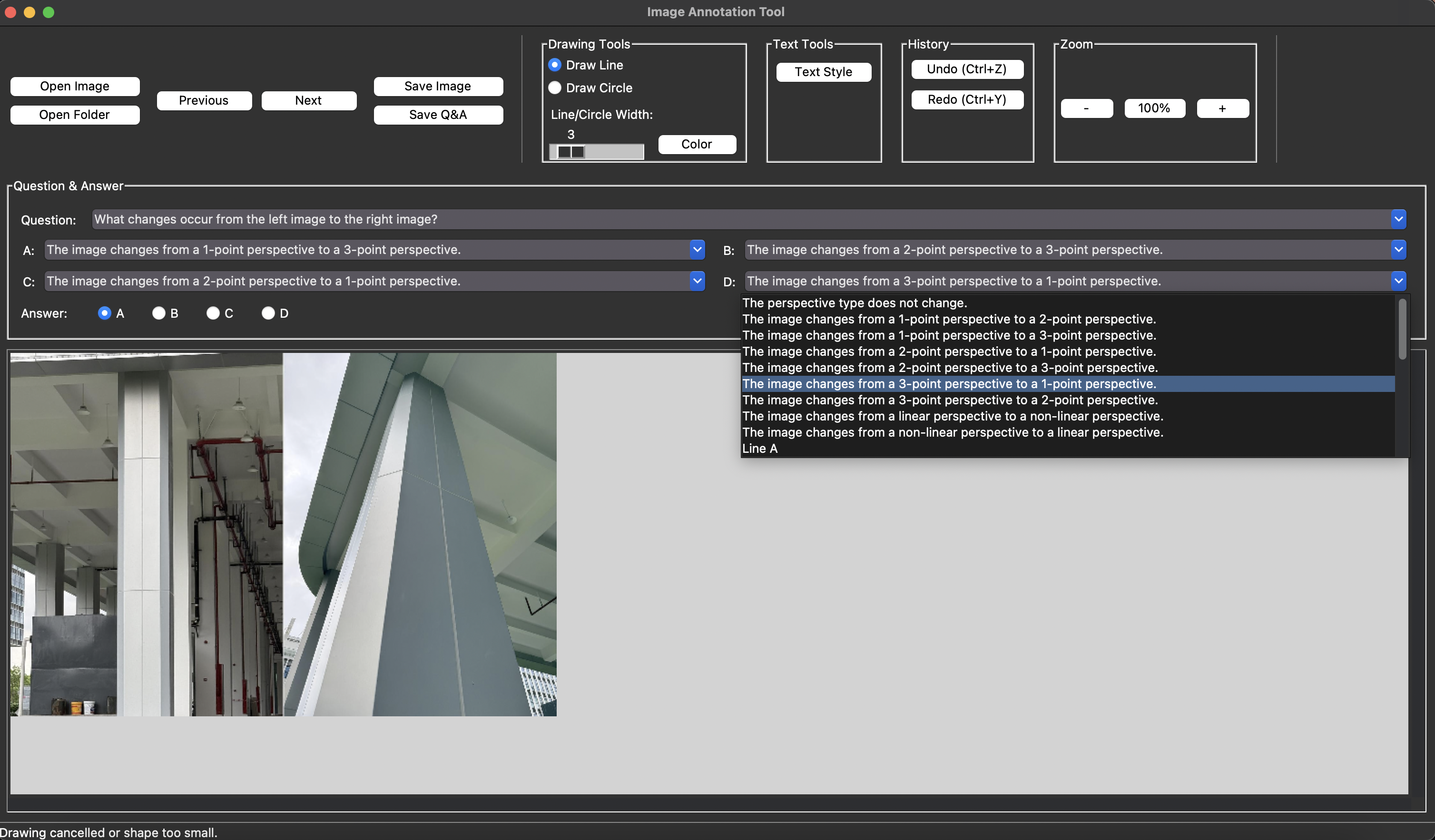}
    \caption{Annotation interface developed for constructing perspective-based multiple-choice questions. The tool integrates geometric drawing utilities, structured answer selection, and image navigation to support precise and consistent labeling.}
    \label{fig:anno_tool}
\end{figure}

We develop a dedicated annotation tool (see \Cref{fig:anno_tool}) to support the systematic construction of multiple-choice questions in our benchmark. Designed specifically for perspective understanding, the tool enables annotators to load image pairs, formulate perspective-related questions, and select answers from a predefined list of geometric transformations (e.g., “1-point to 3-point perspective”, “2-point to 1-point perspective”). This standardization ensures consistent labeling across the dataset.
The interface integrates a suite of carefully designed features to facilitate precise annotation. Annotators can draw lines and circles to mark vanishing directions, orthogonal structures, or other relevant cues. Adjustable line width, zoom controls, and undo/redo functionality support detailed inspection and flexible editing. The tool also provides step-wise navigation through image sets and supports saving both visual annotations and structured Q\&A data.
By tailoring the design to the specific needs of perspective-based reasoning, the tool enables the efficient generation of high-quality, semantically grounded tasks. It plays a central role in ensuring the accuracy, consistency, and scalability of our benchmark construction.

\subsection{Annotator Background and Expertise}
To further clarify, our annotation process involved a valuable collaboration between our internal research team and an external team of domain experts. Our internal team consists of 12 graduate students with backgrounds in computer vision, who received specific training on perspective principles using our custom annotation tool. Additionally, we partnered with a firm specializing in artistic perspective training. Their team of professional instructors, after learning about our project, generously contributed a portion of highly specialized annotations on a pro bono basis. This collaboration ensured our benchmark benefits from both technical computer vision oversight and deep, practical expertise in geometric perspective, guaranteeing a high quality of annotation.

\section{More Results}
\label{app:icl}

We also have experiments to assess the effect of In-Context Learning (ICL). Our experiment followed a rigorous one-shot In-Context Learning paradigm. For each test question, we randomly sampled another distinct image-question-answer pair from the same task category and prepended it to the prompt as an example. This approach primes the model with the task format without leaking the test answer. The results are shown in \Cref{tab:icl}.

\begin{table}[!ht]
\caption{The experiments to assess the effect of in-context learning (ICL).}
\label{tab:icl}
\vspace{2mm}
\centering
\resizebox{\textwidth}{!}{
\begin{tabular}{l|cccc|ccccc|ccc|cc}
\toprule
& \multicolumn{4}{c|}{\textbf{Perspective Perception}} 
& \multicolumn{5}{c|}{\textbf{Perspective Reasoning}} 
& \multicolumn{3}{c|}{\textbf{\textcolor{my_red}{P'Percep} \& \textcolor{my_blue}{P'Reason}}} 
& \multicolumn{2}{c}{\textbf{\textcolor{my_green}{Robustness}}} \\
\cmidrule(lr){2-5} \cmidrule(lr){6-10} \cmidrule(lr){11-13} \cmidrule(lr){14-15}
\textbf{Settings} 
& \textbf{\textcolor{my_red}{VPP}} 
& \textbf{\textcolor{my_red}{CLP}} 
& \textbf{\textcolor{my_red}{VAP}} 
& \textbf{\textcolor{my_red}{LDP}} 
& \textbf{\textcolor{my_blue}{PTR}} 
& \textbf{\textcolor{my_blue}{LRR}} 
& \textbf{\textcolor{my_blue}{OVR}} 
& \textbf{\textcolor{my_blue}{PTS}} 
& \textbf{\textcolor{my_blue}{VPC}} 
& \textbf{P Acc} 
& \textbf{R Acc} 
& \textbf{Overall} 
& \textbf{{Graded}} 
& \textbf{{Binary}} \\
\midrule
GPT-4o-mini         & 35.3 & 24.4 & 43.2 & 71.6 & 43.1 & 29.8 & 14.6 & 31.0 & 45.6 & 43.6 & 32.8 & 37.6 & 28.7 & 10.8 \\
GPT-4o-mini (ICL)   & 28.2 & 25.2 & 53.1 & 76.8 & 28.1 & 26.5 & 16.9 & 25.8 & 14.9 & 45.8 & 22.4 & 34.1 & 17.9 & 6.2 \\
GPT-4o              & 42.9 & 35.0 & 66.0 & 86.0 & 82.0 & 41.7 & 29.9 & 33.8 & 32.5 & 57.5 & 44.0 & 50.0 & 71.9 & 49.9 \\
GPT-4o (ICL)        & 55.1 & 43.1 & 71.6 & 92.6 & 80.7 & 51.7 & 40.9 & 42.3 & 39.5 & 65.6 & 51.0 & 58.3 & 72.2 & 53.5 \\
\bottomrule
\end{tabular}
}
\end{table}

Our analysis indicates that ICL's effectiveness may be tied to model scale. The one-shot example significantly increases the performance of the larger GPT-4o (increasing overall accuracy from 50.0\% to 58.3\%), but appears to be detrimental to the smaller GPT-4o-mini (decreasing from 37.6\% to 34.1\%). This divergence suggests that larger models may be better equipped to generalize from in-context examples for this task.

\section{Limitations}
While \NAME provides a comprehensive benchmark for evaluating perspective understanding in MLLMs, several limitations should be acknowledged. First, the benchmark primarily focuses on static images and multiple-choice question answering (MCQA) formats, which may not fully capture the depth of spatial reasoning required in dynamic or open-ended tasks. Real-world applications often demand free-form generation, spatial manipulation, or multi-turn interactions that extend beyond our current evaluation scope. Our choice of the MCQA format was a deliberate design decision based on several considerations:

\begin{itemize}
    \item \textbf{Objectivity and Scalability:} The MCQA format allows for automated, objective, and large-scale evaluation, avoiding the subjectivity and high cost associated with evaluating open-ended responses.

    \item \textbf{Controlled Probing:} This format enables us to precisely probe a model's understanding of specific geometric concepts (e.g., distinguishing between ``two-point'' and ``three-point'' perspective) without the noise from variations in natural language generation.

    \item \textbf{A Foundational First Step:} As the first benchmark in this domain, we believe that establishing a controlled and rigorous evaluation framework is a critical first step. It lays a solid foundation for future work that can build upon our benchmark with more complex, generative tasks.
\end{itemize}

Second, although we curated a diverse set of real and synthetic images, the dataset still exhibits a bias toward architectural and indoor scenes, which may limit generalizability to natural environments or abstract visual contexts. Third, despite our efforts to standardize evaluation, some tasks inevitably contain ambiguous visual cues, and model errors may stem from subjective interpretations rather than a lack of geometric understanding. Lastly, our benchmark assumes that all correct answers are equally accessible across models without considering differences in input modalities, prompting formats, or underlying vision-language alignment strategies. Future work could address these limitations by incorporating more open-ended tasks, expanding domain diversity, and developing adaptive evaluation protocols that account for model-specific reasoning pathways.

\end{document}